\documentclass[10pt]{article}

\usepackage{amsmath}
\usepackage{amssymb}
\usepackage{amsthm}
\usepackage{longtable}
\usepackage{array}
\usepackage{float}       
\usepackage{jevbench}     

\newtheorem{definition}{Definition}
\newtheorem{proposition}{Proposition}
\theoremstyle{definition}
\newtheorem{example}{Example}
\theoremstyle{remark}
\newtheorem{remark}{Remark}

\newcommand{\TV}{\operatorname{TV}}
\hypersetup{pdftitle={JevBench: Metamorphic Coherence Testing for Typed Probabilistic Decision Models},
  pdfauthor={Chen Feng}, pdfsubject={Technical report},
  pdfkeywords={metamorphic testing, coherence, typed decision models, benchmark}}

\newcommand{\nModels}{21}
\newcommand{\nFull}{21}
\newcommand{\topModel}{Open-Jev-9B}
\newcommand{\topScore}{83.9}
\newcommand{\bottomScore}{56.5}
\newcommand{\medianScore}{72.2}
\newcommand{\dimREPMean}{74.0}

\newcommand{\dimMEAMean}{36.8}

\newcommand{\dimMEAMax}{52.5}
\newcommand{\dimLOGMean}{75.1}

\newcommand{\dimCHOMean}{75.5}

\newcommand{\nBATperfect}{19}
\newcommand{\agreeN}{21}
\newcommand{\agreePearson}{0.992}
\newcommand{\agreeSpearman}{0.983}
\newcommand{\agreeKendall}{0.914}
\newcommand{\agreeMeanDiff}{+0.55}

\newcommand{\agreeMaxAbsDiff}{2.17}
\newcommand{\agreeMaxRankShift}{2}
\newcommand{\agreePairs}{210}
\newcommand{\agreePairsSame}{201}
\newcommand{\agreeSep}{187}
\newcommand{\agreeSepSame}{186}
\newcommand{\agreeBATSpearman}{0.84}
\newcommand{\ciWidthMini}{4.4}
\newcommand{\ciWidthFull}{1.7}
\newcommand{\ciRatio}{2.5}
\newcommand{\hardestRelations}{\texttt{score\_cumulative} (1), \texttt{choice\_complement} (6), \texttt{score\_complement} (6), \texttt{negation\_wrapper} (7), \texttt{choice\_indicator} (8)}

\newcommand{\sizeSpearman}{0.48}

\newcommand{\fastestSecPerReq}{0.037}
\newcommand{\slowestSecPerReq}{6.11}

\newcommand{\baseFakeBiased}{58.6}
\newcommand{\baseFakeCoherent}{64.9}
\newcommand{\baseRandom}{34.5}
\newcommand{\baseUniform}{89.4}
\newcommand{\negTemplatedViolations}{468}
\newcommand{\negTemplatedSameVerdict}{87}
\newcommand{\negTemplatedMean}{7}

\newcommand{\negNaturalSameVerdict}{51}
\newcommand{\negNaturalMean}{42}
\newcommand{\accMin}{55.8}
\newcommand{\accMax}{89.2}
\newcommand{\accN}{1938}
\newcommand{\accCohSpearman}{0.64}
\newcommand{\accDecoderMin}{82.4}
\newcommand{\accDecoderMax}{89.2}
\newcommand{\cohDecoderMin}{64.2}
\newcommand{\cohDecoderMax}{83.9}
\newcommand{\nDecoder}{14}
\newcommand{\brierMin}{0.145}
\newcommand{\brierMax}{0.541}
\newcommand{\brierCohSpearman}{-0.69}
\newcommand{\brierAccSpearman}{-0.97}
\newcommand{\logLossBrierSpearman}{0.98}
\newcommand{\brierUniform}{0.602}
\newcommand{\logLossUniform}{0.976}
\newcommand{\baseAccUniform}{39.8}
\newcommand{\baseAccRandom}{39.6}
\newcommand{\baseAccFakeCoherent}{49.9}
\newcommand{\baseAccFakeBiased}{41.4}
\newcommand{\domDeviation}{1.3}
\newcommand{\domHalfWidth}{2.8}
\newcommand{\domOutside}{8}
\newcommand{\domModels}{21}
\newcommand{\spotChecker}{GPT-5.6 Sol}
\newcommand{\spotN}{48}
\newcommand{\spotCases}{24}
\newcommand{\spotAgree}{85.4}
\newcommand{\spotAgreeLo}{72.8}
\newcommand{\spotAgreeHi}{92.8}

\newcommand{\spotNoulN}{23/24}

\newcommand{\spotChoiceN}{12/12}

\newcommand{\spotScoreN}{6/12}
\newcommand{\spotRewriteN}{40}
\newcommand{\spotRewriteValid}{39}
\newcommand{\spotDisagree}{7}

\newcommand{\menuIndTests}{504}
\newcommand{\menuIndFailed}{464}
\newcommand{\menuIndUnexplained}{429}
\newcommand{\menuIndUnexplainedPct}{92}
\newcommand{\menuCompTests}{504}
\newcommand{\menuCompFailed}{476}
\newcommand{\menuCompUnexplained}{406}
\newcommand{\menuCompUnexplainedPct}{85}
\newcommand{\menuCloTests}{504}
\newcommand{\menuCloFailed}{272}
\newcommand{\menuCloUnexplained}{111}
\newcommand{\menuCloUnexplainedPct}{41}

\newcommand{\wEqualSpearman}{0.94}
\newcommand{\wDropMin}{0.92}
\newcommand{\wDropMax}{0.99}
\newcommand{\wTopUnchanged}{unchanged}
\newcommand{\splitN}{21}
\newcommand{\splitTests}{10{,}800}
\newcommand{\splitSpearman}{0.983}
\newcommand{\splitPearson}{0.990}
\newcommand{\pairsTotal}{210}
\newcommand{\pairsPaired}{172}
\newcommand{\pairsOverlap}{153}
\newcommand{\pairTopDiff}{2.7}
\newcommand{\pairTopLo}{0.7}
\newcommand{\pairTopHi}{4.8}
\newcommand{\pairTopSeparated}{20}
\newcommand{\vacAndLower}{84}
\newcommand{\vacOrUpper}{76}
\newcommand{\vacCountLower}{22}
\newcommand{\vacCountUpper}{13}
\newcommand{\tolBandSpearmanMin}{0.95}
\newcommand{\tolBandShiftMax}{4}

\newcommand{\tolLooseSpearman}{0.60}
\newcommand{\tolLooseMean}{87.2}
\newcommand{\tolStrictMean}{41.4}
\newcommand{\tolUniformLoose}{95.3}
\newcommand{\tolAverageSpearman}{0.98}
\newcommand{\tolClmStrictGain}{14}
\newcommand{\tolVerdictLooseGain}{15}
\newcommand{\gpuHours}{38.2}
\newcommand{\cpuHours}{28.2}

\newcommand{\nCases}{240}
\newcommand{\nDomains}{12}
\newcommand{\nQuestions}{2{,}240}
\newcommand{\nNoul}{1{,}341}
\newcommand{\nChoice}{615}
\newcommand{\nScore}{284}

\newcommand{\qPerCaseMin}{8}
\newcommand{\qPerCaseMax}{12}
\newcommand{\wordsMean}{183}
\newcommand{\wordsMin}{66}
\newcommand{\wordsMax}{382}
\newcommand{\nGold}{1{,}938}
\newcommand{\nUnknown}{302}
\newcommand{\pctUnknown}{13.5}
\newcommand{\nRewriteCandidates}{1{,}697}
\newcommand{\nRewriteJudgeRejected}{529}
\newcommand{\nRewriteReviewRejected}{345}
\newcommand{\nRewriteUsed}{823}
\newcommand{\nRewriteSources}{441}
\newcommand{\nPlanTests}{12{,}000}
\newcommand{\nPlanRequests}{10{,}320}
\newcommand{\nPlanCandidates}{83{,}964}
\newcommand{\nMiniTests}{1{,}200}
\newcommand{\nMiniRequests}{1{,}248}
\newcommand{\nRewriteParaphrase}{294}
\newcommand{\nRewriteChinese}{180}
\newcommand{\nRewriteNegation}{168}
\newcommand{\nRewriteDescription}{181}

\title{JevBench: Metamorphic Coherence Testing for Typed Probabilistic Decision Models}
\author{Chen Feng}
\affiliation{ML Lab, Queen's University Belfast}
\email{c.feng@qub.ac.uk}
\website{jevbench.github.io}
\repository{github.com/JevBench/jevbench}
\package{pypi.org/project/jevbench}

\begin{document}
\maketitle

\begin{abstract}
Typed probabilistic decision models return distributions over specified
answers, such as yes or no, a set of options or levels on an ordered
scale. We ask whether their probabilities remain coherent when
questions about the same input are reworded, negated, combined, given
different options or placed in different batches. JevBench tests this
coherence without gold labels. Its 50 metamorphic relations each test
one law of probability or choice across five dimensions:
representation, batch independence, probability measure, logical order
and choice-set conditioning. Each test changes the question or batch
while keeping the input fixed. Every model receives the same requests,
fixed in advance. JevBench-240 contains \nPlanTests{} tests across
\nCases{} synthetic cases in \nDomains{} decision domains. A fixed
subset of \nMiniTests{} tests, JevBench-mini, closely reproduces the
full-suite ranking (Spearman \agreeSpearman). Across \nModels{} openly
released models, coherence scores range from \bottomScore{} to
\topScore{}. Batch independence holds for most models. Additivity
fails broadly, particularly for complements, where the probabilities
of a proposition and its negation should sum to one. A uniform
reference model ignores the input and scores \baseUniform{} at chance
accuracy, exceeding every evaluated model's coherence score. Among
decoders, accuracy ranges from \accDecoderMin\% to \accDecoderMax\%,
while coherence ranges from \cohDecoderMin{} to \cohDecoderMax{}.
These results show why both measures are needed to assess decision
models. JevBench is available as an open-source Python package.
\end{abstract}


\section{Introduction}
\label{sec:introduction}

A growing class of small models answers typed questions about text.
For example, a model can decide whether a message requests a refund,
which team should handle a ticket or how urgent an incident is on a
four-level scale. It returns a probability distribution in each case:
over yes and no (a \emph{Noul}), named options (a \emph{Choice}) or
levels on an ordered scale (a \emph{Score}). The
models differ widely in how they are built (encoders that score one
marker per option, fine-tuned decoders that read the logits of answer
letters, pointer and scorer heads on decoder backbones, contrastive
embeddings), but many of them now implement the same wire interface,
which we call Jev: a request carries a fixed \emph{state} and a batch
of typed questions, and the response returns one distribution per
question. Such models inherit the progress of representation learning
\cite{Feng_2022_ASCL,Feng_2023_CVPR,gao2024hypercolumn,Sun_2024_CVPR}
and join a wider use of learned predictors across vision, language and
medicine
\cite{feng2025gentrack,ge2026viewsam,mamo2026agraphhf,masoud2026beyond}.\footnote{The interface is a single HTTP endpoint,
\texttt{POST /v1/systemone}. JevBench is not affiliated with any
provider of the interface; ``Jev'' only names it.}

A model can answer individual questions accurately and still make
inconsistent decisions when a system combines its answers. The
probabilities of a proposition and its negation should sum to one.
Merging two options should add their probabilities, and a conjunction
should be no more probable than either conjunct. Removing an irrelevant
option should preserve the ranking of the remaining options. A
question's answer should also be independent of the other questions in
its batch. These properties matter when a routing system compares team
probabilities, a moderation pipeline applies a probability threshold,
or a workflow negates a condition, merges options or removes an
inapplicable option.

Accuracy benchmarks score each answer against a gold label.
Coherence tests instead check whether distributions for related
questions satisfy a law of probability or choice. The law specifies a
relationship between outputs independently of the correct answers.
Metamorphic testing uses this principle: it transforms an input and
checks whether the resulting outputs satisfy the expected relationship
\cite{chen2018mt,segura2016survey}.

We introduce JevBench, a metamorphic coherence benchmark for typed
probabilistic decision models. Its contributions are:
\begin{itemize}
    \item \textbf{A formal definition} of a metamorphic relation for
    typed probabilistic outputs, with three forms of oracle
    (distributional equality, identity, inequality) and one violation
    measure for each, all in probability units, so that one tolerance
    can be used for every relation (\S\ref{sec:method}).
    \item \textbf{A taxonomy of 50 relations} in five dimensions,
    classified by the operation that links the two compared
    distributions: representation (REP), batch independence (BAT),
    probability measure (MEA), logical order (LOG) and choice-set
    conditioning (CHO). Each relation states exactly one law, and we
    kept a relation only if we found that its law does not follow from
    the others' (\S\ref{sec:families}, Appendix~\ref{app:excluded}).
    \item \textbf{A frozen benchmark} of \nCases{} synthetic cases in
    \nDomains{} decision domains, generated fact-first so that every
    question has a rule-based gold annotation (an explicit unknown when
    the facts it needs are absent), with every test and every request
    fixed in advance: JevBench-240, with \nPlanTests{} tests, and
    JevBench-mini, a fixed subset of \nMiniTests{} tests that we report
    (\S\ref{sec:benchmark}).
    \item \textbf{An evaluation of \nModels{} openly released models}
    from seven architecture families, served with their developers'
    own code, together with reference models and the accuracy of every
    model's unmodified answers (\S\ref{sec:experiments}).
    \item \textbf{An open-source package},
    \texttt{pip install jevbench}, that runs the benchmark against any
    server implementing the interface or any Python function, with a
    request cache and replicas.
\end{itemize}

Overall coherence ranges from \bottomScore{} to \topScore.
Batch independence holds for \nBATperfect{} of the \nModels{} models.
The two exceptions arise from how the models process a batch
(\S\ref{sec:batch}). Probability-measure coherence is the weakest
dimension for every model, and its complement group, where a question
and its negation should have yes probabilities that sum to one, is the
weakest group. No model scores above 32.5 on it. For a template such as
``Is it false that the answer to the following question is yes?'', the
source and negated questions pass this test in 7\% of cases on average. A uniform reference ignores the input and scores \baseUniform{} at
chance accuracy, exceeding every evaluated model's coherence score (\baseAccUniform\%). Among the \nDecoder{} decoder
models, accuracy lies between \accDecoderMin\% and \accDecoderMax\%,
while coherence ranges from \cohDecoderMin{} to \cohDecoderMax{}.
We therefore report accuracy and coherence together.


\section{Related work}
\label{sec:related}

\paragraph{Metamorphic and behavioural testing.}
Metamorphic testing checks a program against relations between the
outputs of related inputs when no oracle for a single output exists
\cite{chen2018mt,segura2016survey}. It has been applied to machine
learning classifiers, where relations such as permuting the training
data or the attributes must leave predictions unchanged
\cite{xie2011ml}. CheckList \cite{ribeiro2020checklist} brought
behavioural testing to NLP models with invariance and directional
tests built from templates and perturbations. JevBench uses laws of probability and choice to measure the size of
violations in predicted distributions. Its fixed test set gives every
model the same requests.

\paragraph{Consistency of language models.}
Pretrained language models give inconsistent answers to paraphrased
queries \cite{elazar2021consistency} and often fail to change their
answer under negation \cite{kassner2020negated}. Logical constraints
between related questions have been used both to measure and to train
for consistency \cite{li2019logic}, and BECEL collects several
consistency types, including negational, symmetric and transitive
consistency, into one benchmark \cite{jang2022becel}. For
probabilistic predictions, consistency checks such as negation,
monotonicity and Bayes' rule have been proposed to evaluate models
whose correctness cannot be verified directly
\cite{fluri2024superhuman}, and extended to forecasting
\cite{paleka2024forecasters}. JevBench applies the same principle to typed probabilistic decision
models that share an interface. Their full distributions over specified
answers allow direct tests of additivity, menu conditioning and batch
independence.

\begin{table*}[t]
\centering
\caption{Comparison of JevBench and related consistency evaluations.
\emph{Gold} indicates whether the consistency score requires correct
answers. BECEL is left unclassified because its test cases differ.
\emph{Test set} describes how the inputs are selected.}
\label{tab:related}
\footnotesize
\renewcommand{\arraystretch}{1.15}
\setlength{\tabcolsep}{3pt}
\begin{tabular}{@{}>{\raggedright\arraybackslash}p{2.4cm}>{\raggedright\arraybackslash}p{3.0cm}>{\raggedright\arraybackslash}p{3.8cm}>{\raggedright\arraybackslash}p{2.2cm}c>{\raggedright\arraybackslash}p{2.4cm}@{}}
\toprule
Work & What is tested & Laws & Answers compared & Gold & Test set\\
\midrule
CheckList \cite{ribeiro2020checklist} & predictions of NLP models on three tasks & capability tests: minimum functionality, invariance, directional expectations & predicted labels & partly & generated per task with a tool\\
ParaRel \cite{elazar2021consistency} & factual knowledge of pretrained LMs & invariance under paraphrase & cloze predictions & no & 328 paraphrases of 38 relations\\
Logic-driven consistency \cite{li2019logic} & natural language inference & logic rules over related examples & predicted labels & no & built from NLI data\\
BECEL \cite{jang2022becel} & pretrained LMs on downstream tasks & several types of behavioural consistency (19 test cases) & task predictions & -- & fixed benchmark\\
Consistency checks \cite{fluri2024superhuman} & chess engines, forecasters, legal judges & logical, human-interpretable rules per task & valuations, forecasts, decisions & no & generated per task\\
Forecaster consistency \cite{paleka2024forecasters} & LLM forecasters & logical relations between questions, scored by arbitrage & forecast probabilities & no & generated question sets\\
\textbf{JevBench} & typed decision models behind one interface & 50 laws of representation, batch independence, additivity, logical order and choice-set conditioning & full distributions over yes/no, options or levels & no & frozen requests, one plan for every model\\
\bottomrule
\end{tabular}
\end{table*}

\paragraph{Positioning.}
Table~\ref{tab:related} places JevBench among these evaluations. Three
choices set it apart from them. First, it compares full distributions
over typed answers, including Choices over several options and ordered
Scores, rather than labels or the probability of a single proposition.
These distributions let us test additivity when options are merged,
split or coarsened, and check whether a Noul agrees with the marginal
probability of the Choice it summarises. They also let us test Luce's
choice axiom when the menu changes \cite{luce1959}.
Second, it treats the request as part of the input: a decision model
answers a batch of questions at once, and batch independence is one of
its five dimensions. Third, every test and every request is frozen in
advance and judged in one unit, probability, at one tolerance, so that
models of any architecture are scored on the same tests and their
scores are comparable. Like the consistency checks of
\cite{fluri2024superhuman,paleka2024forecasters}, JevBench needs no
correct answers; \cite{paleka2024forecasters} found that their
consistency metrics correlate with forecasters' later Brier scores, . We likewise report coherence and accuracy together
(Section~\ref{sec:accuracy}).

\paragraph{Sensitivity to surface form and batching.}
Large language models are sensitive to the order of answer options
\cite{pezeshkpour2024order}, prefer some option identifiers over
others \cite{zheng2024selectors}, and change their accuracy with
semantically irrelevant changes of prompt format
\cite{sclar2024formatting}. Answering several questions in one prompt
reduces cost \cite{cheng2023batch} but lets the questions influence
each other. JevBench's representation and batch dimensions test these
effects as laws: option order and identifiers are relabellings with
known effect, and a question's answer must not depend on its
companions.

\paragraph{Coherence of probability and choice.}
A coherent assignment of probabilities is one that admits no sure
loss, which forces the additivity laws of a probability measure
\cite{definetti1937}; human probability judgments violate the
conjunction rule \cite{tversky1983conjunction}, whose bounds go back
to Fr\'echet \cite{frechet1951}. Luce's choice axiom
\cite{luce1959} requires choice probabilities to be independent of
irrelevant alternatives and implies regularity \cite{block1960}; its
well-known failures for similar alternatives \cite{tversky1972} and
decoys \cite{huber1982,simonson1989} determine which menu changes
JevBench tests with conditioning and which with the partition law
(Appendix~\ref{app:invalid}).

\paragraph{Accuracy and calibration.}
Calibration measures whether predicted probabilities match observed
frequencies \cite{guo2017calibration}, and proper scoring rules such as
the Brier score \cite{brier1950verification} reward accurate
probabilities; holistic evaluations combine many such metrics across
scenarios \cite{liang2023helm}. All of these need correct answers, and
the labels themselves are often noisy
\cite{Feng_2022_BMVC,Feng_2024_NoiseBox,Feng_2024_clipcleaner,feng2025openset,feng2026deconstructing};
evaluations with imperfect judges \cite{feng2025noisy} and certified or
reliability-aware assessments \cite{chen2025prosac,zhi2025tfar} address
parts of this problem.
JevBench's coherence tests require no gold answers. We report
coherence alongside accuracy (Section~\ref{sec:accuracy}).


\section{Method}
\label{sec:method}

JevBench compares a model's outputs for related questions against a
law specified before testing. The fixed state and typed outputs define
what can be compared (Section~\ref{sec:setting}). The relation
specifies the expected constraint (Section~\ref{sec:jevbench_definition}).
The violation measure quantifies departures from that constraint, and
the scores aggregate the results (Sections~\ref{sec:measure} and
\ref{sec:scores}). Sections~\ref{sec:families} and \ref{sec:taxonomy}
group the 50 relations by the constraints they test.

\subsection{Setting}
\label{sec:setting}

Let $\mathcal{S}$ denote the state space and $\mathcal{Q}$ the space
of typed questions. A question $q\in\mathcal{Q}$ consists of a key,
instructions, criteria and an answer type, and is posed alongside the other questions in the request. We treat
those companion questions as part of $q$.
Each question has a finite answer space $\mathcal{Y}_q$: $\{0,1\}$ for
a Noul, the set of option identifiers for a Choice, and the ordered
levels $\{0,\ldots,L-1\}$ for a Score. A Choice forces an answer among
its options: its output is the model's distribution over them given
that the answer is one of them, and a Score likewise over its levels.
For a finite answer space $\mathcal{Y}$,
\begin{equation}
    \Delta(\mathcal{Y})
    =
    \left\{
        P:\mathcal{Y}\rightarrow[0,1]
        \;\middle|\;
        \sum_{y\in\mathcal{Y}}P(y)=1
    \right\}
\end{equation}
is the probability simplex over $\mathcal{Y}$. A Jev model is a typed
probabilistic decision function
\begin{equation}
    f:
    \mathcal{S}\times\mathcal{Q}
    \rightarrow
    \bigcup_{q\in\mathcal{Q}}
    \Delta(\mathcal{Y}_q),
    \qquad
    f(s,q)\in\Delta(\mathcal{Y}_q).
\end{equation}
We write $P_q=f(s,q)$ for the model's answer to $q$ under a state $s$
that is clear from context, and $p_q=P_q(1)$ for the probability of
yes of a Noul.

\subsection{Metamorphic relations}
\label{sec:jevbench_definition}

\begin{definition}[Metamorphic relation]
\label{def:relation}
A JevBench metamorphic relation is a triple
\begin{equation}
    r=\left(A_r,\;T_r,\;\mathcal{R}_r\right),
\end{equation}
where
\begin{enumerate}
    \item $A_r:\mathcal{S}\times\mathcal{Q}\to\{0,1\}$ is the
    \emph{applicability} indicator: $A_r(s,q)=1$ when the construction
    of the relation is valid for the case $(s,q)$;
    \item $T_r$ is a \emph{transformation} that maps a source question
    $q$ to related questions
    \begin{equation}
        T_r(q)=\bigl(q'_1,\ldots,q'_{m_r}\bigr),
    \end{equation}
    all answered under the same state, with the \emph{joint output}
    $\mathbf{P}=\bigl(P_q,P_{q'_1},\ldots,P_{q'_{m_r}}\bigr)$;
    \item $\mathcal{R}_r$ is the \emph{oracle}: a single law that the
    joint output must satisfy, in one of three forms,
    \begin{align}
        &\text{(E) distributional equality:} && \Phi_r(\mathbf{P})=\Psi_r(\mathbf{P}),\label{eq:form_E}\\
        &\text{(I) identity:}                && g_r(\mathbf{P})=0,\label{eq:form_I}\\
        &\text{(L) inequality:}              && g_r(\mathbf{P})\le0.\label{eq:form_L}
    \end{align}
    In (E), $\Phi_r$ and $\Psi_r$ map the joint output to two
    distributions over a common answer space, by pushforward along a
    known map of answers or by renormalisation to a subset of options;
    in (I) and (L), $g_r$ is a real-valued function of the joint output.
    These maps are determined by the questions alone.
\end{enumerate}
A model $f$ is \emph{consistent} with $r$ at a case $(s,q)$ with
$A_r(s,q)=1$ if its joint output satisfies $\mathcal{R}_r$.
\end{definition}

\paragraph{One relation, one law.}
A relation specifies one law. If two laws use the same transformed
questions, they define two relations with a shared transformation. A
law can apply to each member of a finite set fixed by the question,
such as options, levels or batch sizes. In that case, the law holds
exactly when every instance holds.

Most relations need a single related question ($m_r=1$). Relations
over several propositions use more. Four properties hold by
construction.
\begin{enumerate}
    \item[(P1)] \emph{Fixed state.} All questions of a test are answered
    under the state $s$ of the source case; only the question side is
    transformed.
    \item[(P2)] \emph{No ground truth.} The oracle refers only to the
    model's own outputs, through maps fixed by the questions. It never
    refers to the correct answer of the decision problem, and neither
    does the applicability indicator.
    \item[(P3)] \emph{Model agnosticism.} The definition uses the model
    only through $f$, so it applies unchanged to any architecture that
    implements the interface.
    \item[(P4)] \emph{Fixed test set.} The transformation and the
    applicability indicator depend on the case alone, never on the
    model's outputs, so every model answers exactly the same questions
    and all relation scores are computed on the same tests.
\end{enumerate}

\paragraph{What coherence does not measure.}
By (P2), a relation holds or fails whatever the correct answer is:
JevBench measures whether a model's answers are consistent with one
another, not whether they are right. Coherence is therefore necessary
for a trustworthy decision model but not sufficient. A model that
ignores its input and returns a uniform distribution over each question's
declared answers satisfies every invariance and many
additive laws by construction and scores high, while deciding nothing.
Coherence scores should be read beside an accuracy or calibration
evaluation of the same model, never instead of one.

\paragraph{Examples.}
The following examples, all on a customer-support case, show the
three forms.

\begin{example}[Form (E): option order]
\label{ex:permutation}
Let $q$ be the Choice ``Which team should own this request?'' with
options \texttt{billing}, \texttt{engineering}, \texttt{retention}, and
let $q'=T_r(q)$ list the same options in reverse order under the
neutral identifiers \texttt{x}, \texttt{y}, \texttt{z}. With
$\sigma:\texttt{billing}\mapsto\texttt{z},\ \texttt{engineering}\mapsto\texttt{y},\ \texttt{retention}\mapsto\texttt{x}$,
the oracle is
\begin{equation}
    P_{q'}=\sigma_{\#}P_q,
    \qquad\text{that is}\qquad
    P_{q'}(\sigma(o))=P_q(o)\ \text{for every option } o,
\end{equation}
with $\Phi_r(\mathbf{P})=P_{q'}$ and $\Psi_r(\mathbf{P})=\sigma_{\#}P_q$.
\end{example}

\begin{example}[Form (I): complement]
\label{ex:complement}
Let $q$ be the Noul ``Does the customer ask for a refund?'' and $q'$ its
negation ``Is it the case that the customer does not ask for a
refund?''. The oracle is the complement law
\begin{equation}
    g_r(\mathbf{P})=p_q+p_{q'}-1=0.
\end{equation}
The cumulative relation of a Score is an identity stated for every
level: with $q'_k$ the Noul ``Is the level $k$ or higher?'', the law
is $g_{r,k}(\mathbf{P})=p_{q'_k}-\sum_{j\ge k}P_q(j)=0$ for every
$k\ge1$.
\end{example}

\begin{example}[Form (L): conjunction bound]
\label{ex:frechet}
Let $q$ and $q_B$ be the Nouls ``Is a feature failing?'' ($A$) and
``Did the customer change their configuration?'' ($B$), and let
$q_{A\land B}$ ask whether both hold, so that
$T_r(q)=(q_B,q_{A\land B})$. A conjunction is never more probable than
either conjunct \cite{frechet1951}:
\begin{equation}
    g_r(\mathbf{P})=p_{A\land B}-\min\{p_A,p_B\}\le0.
\end{equation}
One transformation may also carry two laws, and then gives two
relations. Duplicating an option $o$ as $o'$ requires both that the
two copies together keep the mass of $o$, $\phi_{\#}P_{q'}=P_q$ with
$\phi(o')=o$, and that they share it equally, $P_{q'}(o)=P_{q'}(o')$.
\end{example}

\subsection{Measuring violations}
\label{sec:measure}

The oracle specifies an exact constraint on real-valued
probabilities. JevBench measures how far the outputs depart from that
constraint, distinguishing, for example, a deviation of $10^{-6}$
from one of $0.5$. The form of the oracle determines the measure.

\begin{definition}[Violation measure]
\label{def:measure}
The violation measure of relation $r$ is
\begin{equation}
    \delta_r(\mathbf{P})
    =
    \begin{cases}
        \TV\bigl(\Phi_r(\mathbf{P}),\Psi_r(\mathbf{P})\bigr) & \text{for form (E)},\\[2pt]
        \bigl|g_r(\mathbf{P})\bigr|                         & \text{for form (I)},\\[2pt]
        \max\bigl\{0,\,g_r(\mathbf{P})\bigr\}               & \text{for form (L)},
    \end{cases}
    \label{eq:measure}
\end{equation}
where
$\TV(P,Q)=\frac{1}{2}\sum_{y}|P(y)-Q(y)|$ is the total variation
distance.
\end{definition}

\noindent
For a law stated over an index, the measure is the largest violation
among its instances,
\begin{equation}
    \delta_r(\mathbf{P})=\max_i\delta_{r,i}(\mathbf{P}),
    \label{eq:measure_index}
\end{equation}
with each $\delta_{r,i}$ given by \eqref{eq:measure}; for the
cumulative law of Example~\ref{ex:complement},
$\delta_r=\max_k|g_{r,k}(\mathbf{P})|$.

\begin{proposition}
\label{prop:zero}
For every relation, $\delta_r(\mathbf{P})\ge0$, and
\begin{equation}
    \boxed{\;
        \delta_r(\mathbf{P})=0
        \iff
        \mathbf{P}\ \text{satisfies}\ \mathcal{R}_r
    \;}.
    \label{eq:zero_set}
\end{equation}
\end{proposition}

\begin{proof}
Each case of \eqref{eq:measure} is nonnegative. $\TV(P,Q)=0$ if and
only if $P=Q$; $|g|=0$ if and only if $g=0$; and $\max\{0,g\}=0$ if and
only if $g\le0$. For a law over an index, the maximum
\eqref{eq:measure_index} of nonnegative terms is zero exactly when
every term is, that is, when every instance holds.
\end{proof}

\noindent
All three measures express deviations in probability units. For (E),
$\TV(P,Q)=\max_{E\subseteq\mathcal{Y}}|P(E)-Q(E)|$ gives the largest
difference between the two distributions on any event. We use one
tolerance for all relations, although the maximum deviation varies by
law. For example, inclusion--exclusion can deviate by up to 2, and the
complement law by up to 1.
Section~\ref{sec:tolerance} reports the scores at other tolerances.

\begin{example}[Measuring the examples]
\label{ex:measure}
Suppose the model answers the three examples as follows.
\begin{itemize}
    \item Example~\ref{ex:permutation}: $P_q=(0.70,0.20,0.10)$ for
    \texttt{billing}, \texttt{engineering}, \texttt{retention}, and
    $0.62$, $0.24$, $0.14$ for their images \texttt{z}, \texttt{y},
    \texttt{x}. Then
    $\delta_r=\tfrac12(0.08+0.04+0.04)=0.08$.
    \item Example~\ref{ex:complement}: $p_q=0.80$ and $p_{q'}=0.25$. Then
    $\delta_r=|0.80+0.25-1|=0.05$.
    \item Example~\ref{ex:frechet}: $p_A=0.60$, $p_B=0.70$,
    $p_{A\land B}=0.66$. Then $\delta_r=\max\{0,\,0.66-0.60\}=0.06$.
\end{itemize}
\end{example}

\paragraph{Tolerance.}
A test $(s,q)$ with $A_r(s,q)=1$ \emph{passes} relation $r$ at
tolerance $\varepsilon\ge0$ if
\begin{equation}
    \boxed{\;
        \delta_r(\mathbf{P})\le\varepsilon
    \;},
    \label{eq:pass}
\end{equation}
and its violation indicator is
$V_r(s,q;\varepsilon)=\mathbb{I}\bigl[\delta_r(\mathbf{P})>\varepsilon\bigr]$.
We use $\varepsilon=0.05$ for every relation and model.
In Example~\ref{ex:measure}, the permuted Choice ($0.08$) and the
conjunction bound ($0.06$) exceed this tolerance. The complement
($0.05$) is on the threshold and passes.

\begin{proposition}
\label{prop:relaxation}
For every relation and test,
\begin{enumerate}
    \item the test passes at $\varepsilon=0$ if and only if the model
    is consistent with $r$;
    \item if the test passes at $\varepsilon$, it passes at every
    $\varepsilon'\ge\varepsilon$;
    \item the test passes at every $\varepsilon>0$ if and only if the
    model is consistent with $r$.
\end{enumerate}
\end{proposition}

\begin{proof}
(1) follows from Proposition~\ref{prop:zero}, since $\delta_r\ge0$, and
(2) is immediate. For (3), $\delta_r(\mathbf{P})\le\varepsilon$ for all
$\varepsilon>0$ exactly when $\delta_r(\mathbf{P})=0$.
\end{proof}

\noindent
The tolerance thus relaxes each law monotonically and recovers it
exactly in the limit.

\paragraph{Relation score.}
Let $(s_1,q_1),\ldots,(s_N,q_N)$ be the cases of the benchmark, and let
\begin{equation}
    N_r=\sum_{i=1}^{N}A_r(s_i,q_i)>0
\end{equation}
be the number of tests of relation $r$. The relation score is the pass
rate over these tests,
\begin{equation}
    \boxed{\;
        S_r(\varepsilon)
        =
        1-\frac{1}{N_r}
        \sum_{i=1}^{N}A_r(s_i,q_i)\,V_r(s_i,q_i;\varepsilon)
    \;}
    \in[0,1].
    \label{eq:relation_score}
\end{equation}
By Proposition~\ref{prop:relaxation}, $S_r(\varepsilon)$ is
nondecreasing in $\varepsilon$, and $S_r(0)$ is the fraction of tests
on which the model is exactly consistent with $r$.

\paragraph{The frozen benchmark.}
The benchmark fixes one test for each relation and case in advance.
For $N$ cases, this gives $N$ tests per relation, each with one
follow-up request. When the instances of a law are all
answered in one request (the options of a Choice, the levels or tails
of a Score, the thresholds of a count, the questions of a reordered
batch), the test measures the maximum \eqref{eq:measure_index} over
them. When each instance needs a request of its own (a template, a
removed option, an option pair, a batch size), the test takes one
instance $i$, and measures $\delta_{r,i}$. Within each domain, tests cycle through the instances of each
relation so that they are used as evenly as possible.
Since $\delta_{r,i}\le\delta_r$, a violated test is still a violated
law; for such relations $S_r$ estimates the fraction of (case,
instance) pairs at which the law holds, which is never lower, and may
be higher, than the fraction of cases at which every instance
holds. The tests, their requests and this choice are frozen with the
cases and are the same for every model.

The reported benchmark, JevBench-mini, is a fixed subset of these
tests with the same number of tests for every relation
(\S\ref{sec:plans}).

\paragraph{Summary.}
Definition and measurement combine into the basic JevBench testing
unit,
\begin{equation}
    \boxed{
    q
    \xrightarrow{\;T_r\;}
    (q'_1,\ldots,q'_{m_r})
    \xrightarrow{\;f(s,\cdot)\;}
    \mathbf{P}
    \xrightarrow{\;\delta_r\;}
    \delta_r(\mathbf{P})
    \xrightarrow{\;\le\varepsilon\,?\;}
    V_r
    }
\end{equation}
The relation $(A_r,T_r,\mathcal{R}_r)$ states one law of coherence;
the form of its oracle fixes the measure $\delta_r$; and the tolerance
$\varepsilon$ decides how far from the law a test may be and still
pass.

\subsection{Estimating the source output}
\label{sec:estimation}

The source output $P_q$ is the model's answer to one request of the
case's questions, and each follow-up output $P_{q'}$ is likewise the
answer to one request, so a test compares one answer with one answer.
Under the tested serving configurations, all 22 self-hosted models
returned identical answers to three repetitions of the same request
(repeat noise $0$). For a model that samples, one request gives one
draw of $P_q$. Any resulting variation enters $\delta_r$, and the same
fixed tolerance applies. As a diagnostic outside the benchmark,
JevBench can request a case $R$ times and report the repeat noise
\begin{equation}
    \nu_q=\max_{t\le R}\TV\bigl(P_q^{[t]},\bar P_q\bigr),
    \qquad
    \bar P_q=\frac{1}{R}\sum_{t=1}^{R}P_q^{[t]} .
\end{equation}
Of the five hosted models we tested this way (three requests of each
of 28 questions), three returned identical answers and two showed mean repeat
noise $0.002$ and $0.010$, well under the tolerance.

\subsection{Scores}
\label{sec:scores}

\paragraph{Which relations are scored.}
Let $\mathfrak{R}$ be the set of 50 relations
(\S\ref{sec:taxonomy}). For a given model, a relation enters the
scores only if it has at least $n_{\min}=5$ applicable tests. We write
$\mathfrak{R}^{\ast}\subseteq\mathfrak{R}$ for the scored relations
and $\mathfrak{R}^{\ast}_d$ for those in dimension $d$.

\paragraph{Dimension and overall score.}
A dimension score is the mean of its relation scores,
\begin{equation}
    \boxed{\;
        S_d=\frac{1}{|\mathfrak{R}^{\ast}_d|}\sum_{r\in\mathfrak{R}^{\ast}_d}S_r
    \;}
    \label{eq:dimension_score}
\end{equation}
and the overall score weights the five dimensions
$\mathcal{D}=\{\mathrm{REP},\mathrm{BAT},\mathrm{MEA},\mathrm{LOG},\mathrm{CHO}\}$
equally,
\begin{equation}
    \boxed{\;
        S_{\mathrm{JevBench}}=\frac{1}{|\mathcal{D}|}\sum_{d\in\mathcal{D}}S_d
    \;}.
    \label{eq:overall_score}
\end{equation}
A dimension with no scored relation is reported as missing, and the
overall score of that model is not reported.

\paragraph{Coverage.}
A model may refuse a request or answer it invalidly, for example when
a server limits the number of questions per request. Such tests are not
judged, and a model that skipped the tests it fails would score higher
on the rest. A relation is therefore scored only when at least 95\% of
its planned tests were judged; otherwise the relation, its dimension
and the overall score are reported as missing, to limit the effect of missing answers. All \nModels{} scored models
answered every planned test of both suites, so no relation was dropped.

\begin{remark}
Groups (\S\ref{sec:taxonomy}) serve interpretation and reporting, and
group scores are the mean of their relation scores. They add no
weighting level to \eqref{eq:dimension_score}: regrouping the
relations inside a dimension changes neither $S_d$ nor
$S_{\mathrm{JevBench}}$. Weighting the dimensions equally, rather than
the relations, keeps a dimension's weight independent of how many
relations it happens to contain.
\end{remark}

\paragraph{Uncertainty.}
Instances from one case share a state and are not independent. We
therefore resample cases with replacement (1{,}000 draws, fixed seed),
recompute every score on each draw, and report the 2.5th and 97.5th
percentiles as a 95\% interval.

\begin{tcolorbox}[enhanced,colback=jbtint,colframe=jbtint,boxrule=0pt,arc=2pt,left=10pt,right=10pt,
  top=7pt,bottom=7pt,borderline west={2.5pt}{0pt}{jbink},fontupper=\small]
{\sffamily\bfseries\color{jbink}Scoring at a glance}\par\vspace{3pt}
\begin{enumerate}[leftmargin=1.4em,itemsep=1pt,topsep=2pt]
    \item \emph{Test.} One relation on one case: the source answer
    $P_q$ comes from the case's base request, each follow-up from one
    request; $\delta_r$ is the total variation (form E), $|g_r|$ (I) or
    $\max\{0,g_r\}$ (L), the largest over the instances answered in one
    request.
    \item \emph{Pass.} $\delta_r\le0.05$, the same tolerance for every
    relation and model.
    \item \emph{Relation score} $S_r$: the share of its tests that pass
    (24 per relation on JevBench-mini, 240 on JevBench-240).
    \item \emph{Dimension score} $S_d$: the mean of its relations' scores
    (REP 15, BAT 5, MEA 17, LOG 10, CHO 3 relations); groups only
    report.
    \item \emph{Overall score}: the mean of the five dimension scores.
    \item \emph{Not scored}: a relation with fewer than five tests or
    under 95\% coverage; a dimension or overall score that would need it.
    \item \emph{Interval}: 1{,}000 bootstrap draws of the cases.
\end{enumerate}
\end{tcolorbox}

\paragraph{Graded score.}
The binary score treats all violations above the tolerance equally.
The graded score for a relation is one minus the mean normalised excess,
$\min\{1,\max\{0,\delta_r-\varepsilon\}/(1-\varepsilon)\}$.
We cap the excess at 1 for laws whose violation measure can exceed 1.
This keeps the score in $[0,1]$. We aggregate the graded scores in the
same way as the binary scores.

\subsection{Semantic mappings and pushforward distributions}
\label{sec:pushforward}

Several relations compare distributions over different answer spaces
whose outcomes are related by a known map. For
$\phi:\mathcal{Y}\to\mathcal{Y}'$ and $P\in\Delta(\mathcal{Y})$, the
pushforward is
\begin{equation}
    (\phi_{\#}P)(y')=\sum_{y\in\mathcal{Y}:\,\phi(y)=y'}P(y).
    \label{eq:pushforward}
\end{equation}
A bijective $\phi$ relabels the probability vector; a many-to-one
$\phi$ sums the masses of the outcomes it merges. Permutation,
renaming, answer-type encoding, merging, coarsening and
marginalisation are all instances of \eqref{eq:pushforward}.

\subsection{Oracle families}
\label{sec:families}

JevBench groups relations by the operation used in the oracle
(Table~\ref{tab:operations}). Different edits can belong to the same
dimension if their oracles use the same operation. Edits to the same
part of a question can belong to different dimensions. We write $P_q$ for a source output and $P_{q'}$
for the corresponding follow-up output; for each family we give the
oracle and its violation measure (\S\ref{sec:measure}).

\begin{table}[h]
\centering
\caption{The five dimensions and the operations that define them.}
\label{tab:operations}
\small
\begin{tabular}{@{}l>{\raggedright\arraybackslash}p{5.4cm}c>{\raggedright\arraybackslash}p{5.2cm}@{}}
\toprule
Dimension & Operation linking the two distributions & Form & What changes\\
\midrule
REP & bijection $\sigma$ of answers & E & the description of the same events\\
BAT & identity & E & the other questions of the request\\
MEA & many-to-one map $\phi$ of outcomes, additive identity & E, I & the events of one measure\\
LOG & order & L & the propositions, ordered by entailment\\
CHO & restriction to a sub-menu & E & the menu a Choice is conditioned on\\
\bottomrule
\end{tabular}
\end{table}

\paragraph{REP: representation consistency.}
Two questions describe the same events, and their outputs must
correspond through the bijection $\sigma$ between their answer spaces:
\begin{equation}
    P_{q'}=\sigma_{\#}P_q,
    \qquad
    \delta=\TV\bigl(P_{q'},\sigma_{\#}P_q\bigr).
\end{equation}
The four groups differ only in $\sigma$.
\emph{Invariance}: no answer changes its meaning, so $\sigma=\mathrm{id}$:
a paraphrase, a translation, typographical noise, an added clarification,
a neutral key, a paraphrased option description, and logically
equivalent propositions such as $\neg(A\land B)$ and $\neg A\lor\neg B$.
Paraphrases, translations and natural negations are generated once,
judged, reviewed and frozen with the cases; each test uses exactly one
of them, the first usable rewrite of its source in the frozen bank, so
every test of a relation has the same form.
\emph{Equivariance}: $\sigma$ is a known relabelling of the answers
(options reordered, Score levels reversed with $\sigma(k)=L-1-k$, option
identifiers renamed); a question with two identical options is mapped
to itself by the relabelling $\tau$ that exchanges them, so
$P_{q'}=\tau_{\#}P_{q'}$.
\emph{Binding}: a surface cue (a question key or an option identifier)
suggests a correspondence that differs from the semantic one, and the
output must follow $\sigma$: a negated or swapped key leaves the meaning
intact, so $\sigma=\mathrm{id}$, and exchanging two options'
identifiers or descriptions makes $\sigma$ the transposition that
follows the descriptions.
\emph{Type}: two answer types encode the same random variable, and
$\sigma$ is the encoding, for example $\{0,1\}\to\{\text{no},\text{yes}\}$
between a Noul and a two-option Choice; for binary answers the total
variation is the gap in the probability of yes.

\paragraph{BAT: batch independence.}
A question is posed together with the other questions of its case. The
follow-up $q'$ is the same question posed with different companions
(alone, in reversed order, with replaced companions, in batches of
other sizes, or twice), and its output must not change:
\begin{equation}
    P_{q'}=P_q,
    \qquad
    \delta=\TV(P_{q'},P_q).
\end{equation}

\paragraph{MEA: probability measure coherence.}
The questions ask about different events of one probability space, and
the outputs must be compatible with a finitely additive probability
measure $\mu$, with $\mu(\Omega)=1$ and $\mu(\biguplus_iA_i)=\sum_i\mu(A_i)$.
The operation is a many-to-one map of outcomes, or the additive
identity it induces.
\emph{Complement}: for an event $A$ asked as two separate questions,
$\delta=|p_A+p_{A^c}-1|$, where $p$ is the probability of yes.
\emph{Partition}: within one question, the follow-up options coarsen or
refine the source options through a map $\phi$ that sends each fine
outcome to the coarse outcome containing it. For a coarsening (merge,
hierarchy, adjacent Score levels merged)
$\delta=\TV(P_{q'},\phi_{\#}P_q)$; for a refinement
$\delta=\TV(\phi_{\#}P_{q'},P_q)$. Refinements split an option by a
condition, duplicate it exactly or in other words, or add a level
beyond the top of a Score, which refines the open-ended top level, so
that $\phi$ maps the new level to the old top. A split option is a
partition of its event. A duplicate, exact or in other words
(``essentially the same as $b$''), names the same event as the
original; we take a forced choice between two names of one event to
share its mass, equally by symmetry, so the two together must keep the
original's mass. Replacing an option $o$ by a catch-all $c$, ``none of
the other options'', relabels the event $o$ when the answer lies in the
menu, so $\phi(o)=c$ (Appendix~\ref{app:menu}).
\emph{Marginalisation}: a question whose answer is a deterministic
function $\pi$ of other questions' outcomes must receive the pushforward
$\pi_{\#}P$. For the cumulative Noul ``level $k$ or higher'' of a Score,
$\pi_k(y)=\mathbb{I}[y\ge k]$ and
\begin{equation}
    p_{\ge k}=(\pi_{k\#}P_q)(1)=\sum_{j\ge k}P_q(j),
    \qquad
    \delta=\max_k\bigl|p_{\ge k}-\textstyle\sum_{j\ge k}P_q(j)\bigr|.
\end{equation}
For the Noul ``is it $o$?'' of a Choice, $\pi_o(y)=\mathbb{I}[y=o]$, so its
answer must equal $P_q(o)$ for every option, when the answer lies in the
menu. For a four-way Choice over
the pairs of answers to $A$ and $B$, the projections onto each
coordinate must equal the separate answers. And the answers to $A$,
$B$, $A\land B$ and $A\lor B$ must be projections of one joint
distribution of $(A,B)$, which requires inclusion--exclusion,
$p_{A\land B}+p_{A\lor B}=p_A+p_B$.

\paragraph{LOG: logical coherence.}
The questions are propositions ordered by entailment, and their
probabilities must respect the order; every oracle is an inequality of
form (L), measured by its excess.
\emph{Bounds}: a conjunction is never more probable than either conjunct
\cite{frechet1951}, $p_{A\land B}\le\min\{p_A,p_B\}$; with
inclusion--exclusion this gives every Fr\'echet bound of two
propositions. For $n\ge3$ propositions no identity links the
conjunction and the disjunction, and the four bounds
\begin{align}
    \max\Bigl\{0,\sum_ip_{A_i}-(n-1)\Bigr\}&\le p_{\bigwedge_iA_i}\le\min_ip_{A_i},\\
    \max_ip_{A_i}&\le p_{\bigvee_iA_i}\le\min\Bigl\{1,\sum_ip_{A_i}\Bigr\}
\end{align}
are four laws. For ``at least two of $A_1,\ldots,A_n$'' with $n\ge3$,
the counting bounds
$\frac{1}{n-1}(\sum_ip_{A_i}-1)\le p_{\ge2}\le\frac12\sum_ip_{A_i}$ follow
from $\mathbb{E}[N]=\sum_ip_{A_i}$ for the number $N$ of true
propositions.
\emph{Monotonicity}: for a chain $A_1\models A_2\models\cdots\models A_n$
(a stricter threshold, a stronger claim, a stronger quantifier),
\begin{equation}
    p_{A_1}\le p_{A_2}\le\cdots\le p_{A_n},
    \qquad
    \delta=\max_{i<j}\max\{0,\,p_{A_i}-p_{A_j}\}.
\end{equation}

\paragraph{CHO: choice-set coherence.}
A Choice is the model's distribution over its options given that the
answer is one of them (\S\ref{sec:setting}). Changing the menu from $C$
to $C'$ therefore changes the event conditioned on. With $K=C\cap C'$
the options the two menus share and $P|_K$ a distribution renormalised
to $K$, conditioning on $K$ within either menu must give the same
result:
\begin{equation}
    P_{q'}|_K=P_q|_K,
    \qquad
    \delta=\TV\bigl(P_{q'}|_K,P_q|_K\bigr).
\end{equation}
This is Luce's independence of irrelevant alternatives
\cite{luce1959}, which implies regularity \cite{block1960}. As a model
of preference it is an assumption, known to fail for similar options
and decoys \cite{tversky1972,huber1982}; here it follows from reading
the outputs as one measure over events conditioned on the menu;
comparing
renormalised distributions keeps $\delta$ bounded. It is tested when an
off-topic option is added, when each option in turn is removed, and
when each pair of options is asked alone, the binary form of the choice
axiom. The law needs the added or removed options to be events
disjoint from the others. An option that overlaps an existing one,
such as a near-duplicate or a level beyond the top of a Score, refines
that option instead, and its law is the partition law of MEA.

\paragraph{Boundaries.}
REP compares two descriptions of the same events; MEA compares
different events of one measure under the same menu; CHO compares the
same events under different menus; LOG orders probabilities rather
than equating them; and BAT keeps the question fixed and changes only
its companions.

\subsection{Taxonomy of the relations}
\label{sec:taxonomy}

Table~\ref{tab:taxonomy} lists the 50 relations by dimension and
group, each with its transformation, its law and the form of the law,
which fixes its violation measure (Definition~\ref{def:measure}). Each
relation is placed by the operation of its oracle, so its position
follows from the definition of its dimension. We fixed the taxonomy before examining model outputs. We excluded
candidate laws that we found were implied by other laws on the same
outputs. This check covers the candidates we considered. The taxonomy
does not establish formal independence. Appendix~\ref{app:excluded}
lists the excluded candidates and the reasons for excluding them.

{\footnotesize
\setlength{\tabcolsep}{4pt}
\begin{longtable}{@{}>{\raggedright\arraybackslash}p{1.8cm}>{\raggedright\arraybackslash}p{4.0cm}c>{\raggedright\arraybackslash}p{3.7cm}>{\raggedright\arraybackslash}p{4.5cm}@{}}
\caption{The 50 relations. $P_q$: source output; $P_{q'}$: follow-up
output; $p$: probability of yes of a Noul; $C$: the original options;
$P|_K$: a distribution renormalised to $K$; $\sigma$, $\phi$, $\pi$:
known maps of answers. Form: (E) distributional equality, (I) identity,
(L) inequality; a law with ``for every'' is stated over an index
(\S\ref{sec:jevbench_definition}).}
\label{tab:taxonomy}\\
\toprule
Group & Relation & Form & Transformation & Law\\
\midrule
\endfirsthead
\toprule
Group & Relation & Form & Transformation & Law\\
\midrule
\endhead
\bottomrule
\endfoot
\multicolumn{5}{@{}l}{\textbf{REP} Representation consistency (15)}\\
invariance & instruction\_paraphrase & E & judged paraphrase of the instructions & $P_{q'}=P_q$\\
 & cross\_lingual & E & question translated & $P_{q'}=P_q$\\
 & typo\_noise & E & typos, casing, punctuation & $P_{q'}=P_q$\\
 & verbosity & E & redundant clarification added & $P_{q'}=P_q$\\
 & key\_rename & E & neutral question key & $P_{q'}=P_q$\\
 & description\_paraphrase & E & one option description paraphrased & $P_{q'}=P_q$\\
 & de\_morgan & E & Nouls $q'_1$: $\neg(A\land B)$ and $q'_2$: $\neg A\lor\neg B$ & $P_{q'_1}=P_{q'_2}$\\
equivariance & option\_permutation & E & options reordered; Score levels reversed & $P_{q'}=\sigma_\#P_q$\\
 & option\_rename & E & neutral option identifiers & $P_{q'}=\sigma_\#P_q$\\
 & clone\_symmetry & E & option $o$ duplicated as $o'$ & $P_{q'}=\tau_\#P_{q'}$, $\tau$ swaps $o,o'$\\
binding & key\_instruction\_conflict & E & key negated, or swapped with another question's & $P_{q'}=P_q$\\
 & name\_description\_conflict & E & two options' identifiers exchanged; descriptions keep their positions & $P_{q'}=\sigma_\#P_q$, $\sigma$ follows the descriptions\\
 & description\_swap & E & two options' descriptions exchanged; identifiers keep their positions & $P_{q'}=\sigma_\#P_q$, $\sigma$ follows the descriptions\\
type & noul\_as\_choice & E & Noul asked as a yes/no Choice & $P_{q'}=\sigma_\#P_q$\\
 & noul\_as\_score & E & Noul asked as a two-level Score & $P_{q'}=\sigma_\#P_q$\\
\midrule
\multicolumn{5}{@{}l}{\textbf{BAT} Batch independence (5)}\\
isolation & batch\_solo & E & $q$ asked alone & $P_{q'}=P_q$\\
 & batch\_order & E & batch order reversed & $P_{q'}=P_q$\\
 & duplicate\_question & E & $q$ asked twice in one batch, as $q'_1,q'_2$ & $P_{q'_1}=P_{q'_2}$\\
 & batch\_content & E & companions replaced, batch size kept & $P_{q'}=P_q$\\
 & batch\_size\_sweep & E & batches $q'_n$ of the first 2, 5 and 10 questions & $P_{q'_n}=P_q$ for every $n$\\
\midrule
\multicolumn{5}{@{}l}{\textbf{MEA} Probability measure coherence (17)}\\
complement & negation\_wrapper & I & templated negation $q'$ of $q$ & $p_q+p_{q'}-1=0$\\
 & natural\_negation & I & natural opposite $q'$ of $q$ & $p_q+p_{q'}-1=0$\\
 & choice\_complement & I & Noul $q'$: ``not $o$'' & $p_{q'}+P_q(o)-1=0$\\
 & score\_complement & I & Noul $q'$: ``not level $k$'' & $p_{q'}+P_q(k)-1=0$\\
 & compound\_negation & I & Nouls $A\land B$ and $\neg(A\land B)$ & $p_{\neg(A\land B)}+p_{A\land B}-1=0$\\
partition & option\_merge & E & options $a,b$ merged & $P_{q'}=\phi_\#P_q$\\
 & option\_split & E & option $o$ split by a condition $X$ & $\phi_\#P_{q'}=P_q$\\
 & option\_clone & E & option $o$ duplicated & $\phi_\#P_{q'}=P_q$\\
 & similarity\_effect & E & near-duplicate $b'$ of $b$ added, in other words & $\phi_\#P_{q'}=P_q$, $\phi(b')=b$\\
 & partition\_hierarchy & E & options grouped into coarse blocks & $P_{q'}=\phi_\#P_q$\\
 & score\_granularity & E & adjacent Score levels merged & $P_{q'}=\phi_\#P_q$\\
 & extreme\_level & E & level beyond the top of a Score added & $\phi_\#P_{q'}=P_q$, $\phi$ maps the new level to the top\\
 & option\_closure & E & each option $o$ in turn replaced by a catch-all $c$, ``none of the other options'' & $P_{q'_o}=\phi_{o\#}P_q$, $\phi_o(o)=c$, for every $o$\\
marginal\-isation & score\_cumulative & I & Nouls $q'_k$: ``level $k$ or higher'' & $p_{q'_k}-\sum_{j\ge k}P_q(j)=0$ for every $k$\\
 & choice\_indicator & I & one Noul $q'_o$: ``is it $o$?'' per option & $p_{q'_o}-P_q(o)=0$ for every $o$\\
 & pair\_choice\_marginals & I & four-way Choice $q'$ over the answers to $A,B$ & $\pi_{i\#}P_{q'}(1)-p_{i}=0$ for every $i\in\{A,B\}$\\
 & inclusion\_exclusion & I & Nouls $A\land B$ and $A\lor B$ & $p_{A\land B}+p_{A\lor B}-p_A-p_B=0$\\
\midrule
\multicolumn{5}{@{}l}{\textbf{LOG} Logical coherence (10)}\\
bounds & frechet\_and\_upper & L & Noul $A\land B$ & $p_{A\land B}-\min\{p_A,p_B\}\le0$\\
 & nary\_and\_upper & L & Noul $\bigwedge_iA_i$, $n\ge3$ & $p_{\bigwedge}-\min_ip_{A_i}\le0$\\
 & nary\_and\_lower & L & Noul $\bigwedge_iA_i$, $n\ge3$ & $\sum_ip_{A_i}-(n-1)-p_{\bigwedge}\le0$\\
 & nary\_or\_upper & L & Noul $\bigvee_iA_i$, $n\ge3$ & $p_{\bigvee}-\sum_ip_{A_i}\le0$\\
 & nary\_or\_lower & L & Noul $\bigvee_iA_i$, $n\ge3$ & $\max_ip_{A_i}-p_{\bigvee}\le0$\\
 & count\_upper & L & Noul ``at least two of $A_1,\ldots,A_n$'', $n\ge3$ & $p_{\ge2}-\tfrac12\sum_ip_{A_i}\le0$\\
 & count\_lower & L & Noul ``at least two of $A_1,\ldots,A_n$'', $n\ge3$ & $\frac{\sum_ip_{A_i}-1}{n-1}-p_{\ge2}\le0$\\
monoton\-icity & threshold\_sweep & L & Nouls ``more than $k$'' for a grid of $k$ & $p_{>k'}-p_{>k}\le0$ for every $k<k'$\\
 & entailment\_strength & L & a claim with a stronger and a weaker version & $p_{A}-p_{B}\le0$ for every $A\models B$\\
 & quantifier\_monotonicity & L & ``all'', ``at least two'', ``at least one'' & $p_{A}-p_{B}\le0$ for every $A\models B$\\
\midrule
\multicolumn{5}{@{}l}{\textbf{CHO} Choice-set coherence (3)}\\
conditioning & irrelevant\_option & E & off-topic option $z$ added & $P_{q'}|_C=P_q$\\
 & remove\_option\_iia & E & each option $o$ in turn removed & $P_{q'_o}=P_q|_{C\setminus\{o\}}$ for every $o$\\
 & pairwise\_iia & E & each pair $\{a,b\}$ of options asked alone, $k\ge4$ options & $P_{q'_{ab}}=P_q|_{\{a,b\}}$ for every pair\\
\end{longtable}
}

\paragraph{Dependence on answer-type equivalence.}
Four MEA relations compare a Noul with a Choice or Score answer:
choice\_complement, score\_complement, score\_cumulative and
choice\_indicator. They assume, beyond additivity, that the two answer types
encode the same event, the property REP.type tests. A model that fails
REP.type will tend to fail these relations as well; we report them in
MEA because the constraint they check is additive. Three relations also
assume that the answer lies in the menu: choice\_indicator and
choice\_complement, whose Nouls do not list the other options, and
option\_closure. Appendix~\ref{app:menu} tests whether the other
reading explains their failures.


\section{The benchmark}
\label{sec:benchmark}

Each benchmark case consists of a state and a batch of typed
questions about it. The cases and rewrite bank supply the inputs for
the fixed test plans (Sections~\ref{sec:cases}, \ref{sec:bank} and
\ref{sec:plans}). Section~\ref{sec:protocol} describes how to run the
benchmark.

\subsection{Cases}
\label{sec:cases}

JevBench-240 contains \nCases{} synthetic cases, 20 in each of
\nDomains{} decision domains (Table~\ref{tab:domains}): customer
support, content moderation, finance, security and IT operations,
e-commerce returns, recruiting, healthcare administration, legal
contracts, logistics, grading and the logs of AI agents. A case has a
state of \wordsMin--\wordsMax{} words (\wordsMean{} on average) and
\qPerCaseMin--\qPerCaseMax{} questions about it, \nQuestions{} in
all: \nNoul{} Nouls, \nChoice{} Choices (four options each) and
\nScore{} Scores (four or five levels).

\begin{table}[t]
\centering
\caption{The twelve domains of JevBench-240: distinct question keys,
questions by type over the domain's 20 cases, and the mean length of
a state.}
\label{tab:domains}
\small
\begin{tabular}{@{}lrrrrr@{}}
\toprule
Domain & Question keys & Nouls & Choices & Scores & Words per state\\
\midrule
Customer support & 14 & 160 & 60 & 20 & 154\\
Content moderation & 15 & 160 & 40 & 40 & 198\\
Finance operations & 19 & 110 & 46 & 34 & 200\\
Security operations & 17 & 115 & 60 & 30 & 195\\
IT incidents & 16 & 98 & 54 & 20 & 190\\
E-commerce returns & 13 & 114 & 60 & 20 & 156\\
HR recruiting & 16 & 105 & 40 & 20 & 181\\
Healthcare administration & 21 & 95 & 45 & 20 & 205\\
Legal contracts & 19 & 105 & 50 & 20 & 174\\
Logistics & 14 & 90 & 60 & 20 & 191\\
Education grading & 15 & 80 & 60 & 20 & 176\\
AI agent logs & 14 & 109 & 40 & 20 & 171\\
\midrule
All & 193 & 1341 & 615 & 284 & 183\\
\bottomrule
\end{tabular}

\end{table}

We generate cases \emph{fact-first} to balance the data by design
and assign each question a rule-based gold annotation. If a required
fact is absent, the annotation records an explicit unknown. For each
domain, we specify fact slots, four scenario archetypes, question
templates with answer rules, and text styles such as a customer email,
a live chat transcript or a moderation queue entry. Generation has four
steps.
\begin{enumerate}
    \item \emph{Plan} (no model). Facts are sampled for an archetype,
    together with how each is mentioned: stated explicitly, implied by
    context, or absent from the text. Every gold answer is computed
    from the facts. Archetypes, gold answers, the position of the
    correct option of a Choice, styles and three length bands are
    balanced greedily over a large pool of plans.
    \item \emph{Write}. A writer model (gpt-oss-20b \cite{openai2025gptoss}) turns the facts
    into a state in the planned style and length. It never sees the
    questions.
    \item \emph{Verify}. A verifier from another model family
    (Llama-3.1-8B-Instruct \cite{grattafiori2024llama3}) reads only the state and reports each
    planned fact, or that it is not stated. Any mismatch with the plan
    rejects the draft.
    \item \emph{Freeze}. Check lengths and near-duplicates, then keep
    20 cases per domain.
\end{enumerate}
When $X$ is absent, the gold answer to ``does the text say $X$'' is
false. The answer to ``is $X$ true'' is recorded as unknown. Of the \nQuestions{}
gold answers, \nUnknown{} (\pctUnknown\%) are unknown and \nGold{} are
known. Neither the writer nor the verifier belongs to a model family
under test, and the gold answers and facts are stored apart from the
cases: a run reads only the states and the questions. gpt-oss-20b is
released under the Apache-2.0 licence and Llama-3.1-8B-Instruct under
the Llama 3.1 Community License; both permit this use, and no model
weights are redistributed.

The cases are synthetic and in English, and their gold answers are
only as good as the plans and the verifier. Coherence does not depend
on them, since no relation refers to a correct answer; the gold answers
are used only for the accuracy we report beside coherence
(\S\ref{sec:accuracy}).

\paragraph{Spot check.}
An independent model, \spotChecker, answered \spotN{} questions of
\spotCases{} cases drawn from JevBench-mini with a fixed seed (two cases
per domain, a yes/no question and a Choice or Score from each) without
seeing the gold answers, and assessed \spotRewriteN{} natural rewrites used in JevBench-mini
(ten of each of the four kinds). The question sample comprised
\spotNoulN{} yes/no questions, \spotChoiceN{} Choices and
\spotScoreN{} Scores. The model agreed with the gold answers on
\spotAgree\% of questions (95\% interval \spotAgreeLo--\spotAgreeHi\%).
We manually reviewed all \spotDisagree{} disagreements and attributed
them to ambiguity in the questions. We found no definite error in
either answer. Most concern Scores, whose gold level follows from a rule
over the planned facts: when an input of the rule is absent from the
text the gold is unknown, although a reader may still infer a level,
and a reader may weigh cues the rule does not use (an expense claim's
amount, say, where the rule looks only at the receipt). This ambiguity affects the accuracy comparison because the coherence
relations do not use gold answers. Four relations do depend on
rewrites. The spot check judged \spotRewriteValid{} of
\spotRewriteN{} rewrites valid; the exception renders an option that names an action as a noun
phrase in the Chinese translation. It remains in the frozen suites and
could affect the cross-lingual test that uses it. A larger audit stratified by relation would be needed to estimate
error rates for individual rewrite types.

\subsection{The rewrite bank}
\label{sec:bank}

Four relations need natural rather than templated rewrites: a
paraphrase of the instructions (\texttt{instruction\_paraphrase}), a
translation (\texttt{cross\_lingual}), a paraphrase of one option
description (\texttt{description\_paraphrase}) and a natural opposite
of a Noul (\texttt{natural\_negation}). They are generated once and
frozen with the cases. gpt-oss-20b proposed \nRewriteCandidates{} rewrites and served as the
initial judge with two votes per rewrite. It rejected
\nRewriteJudgeRejected{} candidates that failed to preserve meaning or,
for negations, failed to express the exact opposite. A second review
by Claude rejected \nRewriteReviewRejected{} more. The
\nRewriteUsed{} remaining rewrites (\nRewriteParaphrase{} paraphrases,
\nRewriteDescription{} option descriptions, \nRewriteChinese{} Chinese
translations and \nRewriteNegation{} negations) cover
\nRewriteSources{} distinct source texts. Beyond the spot check
(\S\ref{sec:cases}), the bank has not been audited by humans. A test uses exactly one rewrite: the first usable
one for its source in the bank's fixed order, so every test of a
relation has the same form.

\subsection{Frozen test plans}
\label{sec:plans}

For every case and every applicable relation, the engine builds
candidate trials: the transformed questions, grouped into the requests
that will carry them. The engine builds trials deterministically in a \emph{sealed}
context. Any attempt to read a model output during construction raises
an error, so test construction is independent of the model (P4). Before
storing a trial, the engine checks that the questions are well formed,
the keys are unique and the state is unchanged.

From \nPlanCandidates{} candidate trials, the plan of JevBench-240
keeps exactly one test per relation and case, \nPlanTests{} tests in
all. When a relation has several variants for a case (several
templates, several options to remove, several batch sizes), the
variant is not drawn at random: within each domain the tests of a
relation take the variants in turn, so that each is used as evenly as possible
(\S\ref{sec:measure}). The plan stores every request of every trial,
\nPlanRequests{} distinct requests including one base request per case,
so that a run only sends them and judges the answers. It draws no
random numbers, and settings that would change the tests (another
seed, rewrite bank or number of trials) are refused for a bundled
suite.

JevBench-mini, the suite we report, keeps the cases and takes a fixed
subset of these tests: each relation on two cases of every domain,
24 tests per relation and \nMiniTests{} in all, chosen once and frozen.
A run sends \nMiniRequests{} requests, about a ninth of JevBench-240.
Every JevBench-mini test is a JevBench-240 test. Section
\ref{sec:mini} measures how closely the two rank models. Each suite
records the SHA-256 of its files, and a modified suite is refused.

\subsection{Protocol and package}
\label{sec:protocol}

A run sends each case's questions once, unchanged (the base request),
and then every request of the plan. The base request supplies the source outputs $P_q$ for every test of
that case. Each follow-up output comes from one request. We use a
fixed tolerance of $\varepsilon=0.05$, require at least five tests
and at least 95\% coverage per scored relation
(Section~\ref{sec:scores}), and compute intervals from 1{,}000
bootstrap samples of the cases \cite{efron1979bootstrap}.

The benchmark is distributed as a Python package (Apache-2.0) with no
dependency beyond an HTTP client. A model is any server implementing
the interface, or any Python function that maps a state and a batch of
questions to answers:
\begin{tcolorbox}[enhanced,colback=jbtint,colframe=jbtint,boxrule=0pt,arc=2pt,left=8pt,right=8pt,
  top=4pt,bottom=4pt,borderline west={2.5pt}{0pt}{jbink},fontupper=\small]
\begin{verbatim}
import jevbench as jb
model = jb.systemone("http://localhost:8000/v1/systemone", model="my-model")
jb.check(model)              # one request: are the answers well formed?
report = jb.evaluate(model)  # JevBench-mini
print(report.table("dimension"))
\end{verbatim}
\end{tcolorbox}
\noindent
\texttt{jb.check} sends one request and stops with a message that names
the question and the expected format if an answer deviates from the
standard one, before a run starts.
Requests are cached by model name, so a rerun or a second suite sends
only the requests not yet answered. Most servers answer one request at
a time, so a run is sped up by replicas of the model, one per GPU; the
package keeps exactly one request in flight per replica, which
reproduces the sequential answers.


\section{Experiments}
\label{sec:experiments}

\subsection{Models and setup}
\label{sec:models}

We evaluate \nModels{} openly released Jev-compatible models
(Table~\ref{tab:models} in Appendix~\ref{app:models}). They span seven
architecture families, named by how a model reads its answer:
encoders that score one marker per option (Laya, Laya typed-decisions,
Von, Verdict); a bi-encoder and a cross-encoder distilled from a
decoder teacher (jevhome-E, jevhome-ettin-1b); decoders that read the
logits of answer letters (decider-4b and -2b, JevK5, OneJev-4B,
Jebadiah 9B, Bespoke-Nimble-9B, JevAny-Qwen direct-token); decoders
with a pointer head over the options (Kev-4B, TinyJev-4B, JevAny-Qwen
and JevAny-Gemma pointer); decoders with a trained scorer head
(Open-Jev-9B, RSI-Jev 2B, OpenThai-SystemOne); and projection heads
over frozen contrastive embeddings (CLM). Sizes range from 48M to
9.7B parameters.

We installed each model from its developers' release at a pinned
revision. We used one NVIDIA A100 40GB per server, except for the two
jevhome models, which ran on CPU. Requests contain the state and
questions. The server supplies any prompt, and we add neither
instructions nor in-context examples. We used the developers' server
where available. For Bespoke-Nimble-9B, we used an adapter that
reproduces the developers' request compiler. For Verdict, we used a
third-party server that reimplements its inference. A further model, NeoHorse-Jev-4B, was run but is not
scored: its server refuses requests with more questions than its own
limit, so one relation (\texttt{threshold\_sweep}) has no answers, and
by the coverage rule (\S\ref{sec:scores}) it has no overall score.

No self-hosted model varied: three identical requests returned
identical answers for all 22 models under the tested serving
configurations (\S\ref{sec:estimation}).
All results are on JevBench-mini; all \nFull{} models were also run on
JevBench-240 (\S\ref{sec:mini}).

\paragraph{Reference models.}
Four models that need no GPU anchor the scale. \emph{Uniform} gives
every answer the same probability and ignores its input. \emph{Random}
draws a fresh distribution (Dirichlet with unit parameters, a uniform
probability for a Noul) for every distinct state and question, seeded
by their hash, so it is deterministic but its answers to any two
different questions are unrelated. The \emph{Luce toy scorer} scores
each option independently by word overlap with the state, and the
\emph{biased} variant adds a first-option bias, a dependence on the
question key and leakage between the questions of a batch.

\paragraph{Accuracy.}
\label{sec:accuracy_def}
We use each model's base outputs to compute accuracy on the \accN{}
questions with known gold answers (Section~\ref{sec:cases}). A
question earns one point if the gold answer is uniquely most probable
and $1/k$ if it is among $k$ answers tied for highest probability. The
calculation uses existing outputs and requires no further requests.

\subsection{Main results}
\label{sec:results}

\begin{table}[t]
\centering
\caption{JevBench-mini results. Overall coherence (the mean of the five
dimension scores) with its 95\% interval; the rank range lists the
ranks the intervals allow, so models whose ranges overlap are not
separated. Acc.: accuracy of the base answers against the gold answers
(\%); Brier: their Brier score (lower is better, \S\ref{sec:accuracy}).
Below the line, the four reference models.}
\label{tab:leaderboard}
\small
\setlength{\tabcolsep}{3.5pt}
\adjustbox{max width=\linewidth}{\begin{tabular}{@{}rclrccccccrr@{}}
\toprule
Rank & Range & Model & Overall & 95\% CI & REP & BAT & MEA & LOG & CHO & Acc. & Brier\\
\midrule
1 & 1--2 & Open-Jev-9B & \textbf{83.9} & 82.5--85.2 & 81.9 & 100.0 & 50.2 & 87.1 & 100.0 & 88.4 & 0.154\\
2 & 1--7 & JevAny-Gemma-4B (pointer) & \textbf{81.2} & 79.0--83.2 & 81.4 & 100.0 & 52.0 & 83.8 & 88.9 & 87.6 & 0.157\\
3 & 2--7 & JevAny-Qwen-4B (pointer) & \textbf{79.4} & 77.2--81.4 & 81.7 & 100.0 & 52.2 & 78.3 & 84.7 & 88.8 & 0.150\\
4 & 2--8 & decider-4b v2.1 & \textbf{78.4} & 76.0--80.7 & 80.3 & 100.0 & 52.5 & 78.8 & 80.6 & 88.6 & 0.152\\
5 & 2--8 & JevAny-Qwen-4B (direct token) & \textbf{77.8} & 75.4--80.2 & 81.4 & 100.0 & 50.2 & 80.8 & 76.4 & 88.6 & 0.150\\
6 & 2--8 & TinyJev-4B & \textbf{77.3} & 75.4--79.2 & 78.3 & 100.0 & 42.2 & 75.8 & 90.3 & 86.6 & 0.193\\
7 & 2--11 & Jebadiah 9B v2 & \textbf{76.9} & 74.4--79.3 & 79.2 & 100.0 & 48.0 & 83.8 & 73.6 & 89.1 & 0.145\\
8 & 4--10 & Von 1.3 & \textbf{75.9} & 74.7--77.1 & 83.1 & 100.0 & 16.4 & 80.0 & 100.0 & 72.2 & 0.439\\
9 & 7--15 & JevK5 v0.3 & \textbf{72.8} & 70.4--75.3 & 76.1 & 100.0 & 43.6 & 76.2 & 68.1 & 88.1 & 0.162\\
10 & 7--15 & OneJev-4B & \textbf{72.5} & 69.9--74.7 & 79.7 & 100.0 & 43.6 & 71.2 & 68.1 & 89.2 & 0.145\\
11 & 8--15 & RSI-Jev v3.0 2B & \textbf{72.2} & 70.1--74.4 & 73.1 & 100.0 & 32.1 & 73.8 & 81.9 & 85.2 & 0.199\\
12 & 9--15 & Kev-4B & \textbf{71.9} & 69.8--73.9 & 74.4 & 100.0 & 30.1 & 67.5 & 87.5 & 87.9 & 0.164\\
13 & 9--15 & decider-2b v10 & \textbf{71.7} & 69.2--74.4 & 78.6 & 100.0 & 38.0 & 67.1 & 75.0 & 84.9 & 0.198\\
14 & 9--15 & jevhome-E & \textbf{69.6} & 68.1--71.3 & 69.2 & 100.0 & 25.2 & 84.2 & 69.4 & 55.8 & 0.507\\
15 & 9--17 & Bespoke-Nimble-9B$^{1}$ & \textbf{69.2} & 66.5--72.1 & 76.4 & 68.3 & 50.5 & 87.1 & 63.9 & 89.1 & 0.154\\
16 & 15--19 & CLM v0.1 & \textbf{66.8} & 65.5--68.1 & 59.2 & 100.0 & 25.5 & 49.6 & 100.0 & 63.5 & 0.541\\
17 & 15--20 & OpenThai-SystemOne & \textbf{64.2} & 61.2--67.0 & 76.9 & 70.8 & 39.7 & 61.3 & 72.2 & 82.4 & 0.261\\
18 & 16--19 & jevhome-ettin-1b & \textbf{64.1} & 61.8--66.2 & 70.3 & 100.0 & 29.2 & 70.8 & 50.0 & 82.3 & 0.260\\
19 & 16--20 & Laya (typed-decisions) & \textbf{63.9} & 61.4--66.3 & 55.0 & 100.0 & 14.5 & 73.8 & 76.4 & 71.0 & 0.415\\
20 & 18--21 & Verdict 1.4$^{2}$ & \textbf{59.4} & 57.0--61.8 & 66.1 & 100.0 & 19.9 & 77.9 & 33.3 & 64.2 & 0.446\\
21 & 20--21 & Laya (english) & \textbf{56.5} & 53.8--59.2 & 52.5 & 100.0 & 16.4 & 67.9 & 45.8 & 70.8 & 0.390\\
\midrule
 & & \textit{Uniform (ignores the input)} & 89.4 & 89.4--89.4 & 100.0 & 100.0 & 47.1 & 100.0 & 100.0 & 39.8 & 0.602\\
 & & \textit{Random (hash-seeded)} & 34.5 & 33.5--35.5 & 18.6 & 100.0 & 4.9 & 48.8 & 0.0 & 39.6 & 0.753\\
 & & \textit{Luce toy scorer} & 64.9 & 63.8--65.9 & 58.6 & 100.0 & 8.6 & 57.1 & 100.0 & 49.9 & 0.677\\
 & & \textit{Luce toy scorer, biased} & 58.6 & 55.8--61.6 & 53.3 & 78.3 & 18.1 & 79.2 & 63.9 & 41.4 & 0.845\\
\bottomrule
\end{tabular}
}
\par\smallskip
\parbox{0.95\linewidth}{\footnotesize$^{1}$\,Bespoke-Nimble-9B: JevBench shim (Nimble's own Modal server does not run locally); same request mapping as Nimble's compiler.py, PyTorch reference runner, current checkpoint. $^{2}$\,Verdict 1.4: OpenJev server (razorback16/openjev), which reimplements Verdict 1.4's prompt and temperatures; Verdict ships no HTTP server.
}
\end{table}

\begin{tcolorbox}[enhanced,colback=jbtint,colframe=jbtint,boxrule=0pt,arc=2pt,left=10pt,right=10pt,
  top=7pt,bottom=7pt,borderline west={2.5pt}{0pt}{jbcoral},fontupper=\small]
{\sffamily\bfseries\color{jbink}Reading the scores}\par\vspace{3pt}
\begin{itemize}[leftmargin=1.2em,itemsep=1pt,topsep=2pt]
    \item \emph{Coherence is not correctness.} A model that ignores its
    input and answers uniformly scores \baseUniform, above every model
    here, at chance accuracy (\baseAccUniform\%); read every score beside
    the accuracy and the Brier score in the last columns
    (\S\ref{sec:accuracy}).
    \item \emph{Rank uncertainty.} Overlapping rank ranges indicate
    that the interval comparison fails to distinguish those models.
    A difference of a few points can fall within the uncertainty of
    JevBench-mini (Section~\ref{sec:mini}).
    \item \emph{Scores are per law.} A relation score is the share of
    tests whose law holds within $0.05$ in probability; the ranking is
    stable for tolerances from $0.03$ to $0.1$ (\S\ref{sec:tolerance}).
\end{itemize}
\end{tcolorbox}

Table~\ref{tab:leaderboard} lists the results. Overall coherence
ranges from \bottomScore{} to \topScore, with a median of
\medianScore. \topModel{} is first, and its interval separates it from
every model except JevAny-Gemma-4B, which a paired bootstrap separates
too (\S\ref{sec:mini}); the next six models are not separated from one
another. The ranking follows the families only
loosely: decoders with scorer, pointer or letter readouts fill the top
of the table, but the encoder Von is eighth and the decoder
OpenThai-SystemOne seventeenth.

\begin{figure}[t]
\centering
\includegraphics[width=0.8\linewidth]{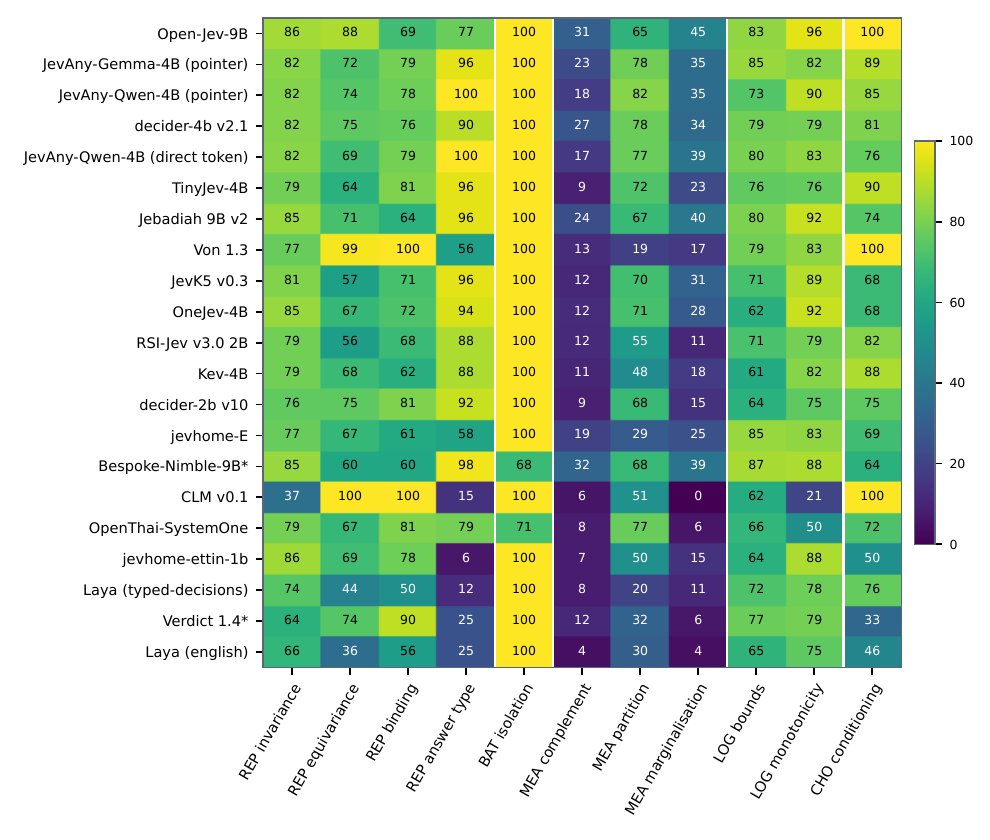}
\caption{Group scores on JevBench-mini, one row per model, ordered by
overall coherence. White lines separate the five dimensions. $^\ast$:
not served by its developers' own server.}
\label{fig:groups}
\end{figure}

\paragraph{Dimensions.}
Figure~\ref{fig:groups} breaks the scores into the eleven groups.
Batch independence (BAT) is nearly saturated: \nBATperfect{} of the
\nModels{} models keep every batch law, and the two that do not are
examined in \S\ref{sec:batch}. Representation (REP, mean
\dimREPMean), logical order (LOG, \dimLOGMean) and choice-set
conditioning (CHO, \dimCHOMean) separate the models. Probability-measure
coherence (MEA) is the lowest dimension of every model, with a mean of
\dimMEAMean{} and no model above \dimMEAMax.

\paragraph{Where models fail.}
The five relations with the lowest mean scores are all in MEA:
\hardestRelations. These relations test the same event across two
answer types or compare a proposition with its negation. A templated negation of a Noul
(for example ``Is it false that the answer to the following question is
yes?'') sums to one with its source in
\negTemplatedMean\% of tests on average, a natural negation from the
rewrite bank in \negNaturalMean\%. In \negTemplatedSameVerdict\% of the \negTemplatedViolations{}
failed templated tests, the yes probabilities for the original and
negated questions are both above or both below one half. This pattern
is consistent with earlier findings for masked language models
\cite{kassner2020negated}. The corresponding share for natural
negations is \negNaturalSameVerdict\%. The cumulative Nouls of a
Score (``level $k$ or higher'') almost never match the Score's own
tail probabilities (1\%), and the Noul ``is it $o$?'' rarely matches
the probability of option $o$ in the Choice (8\%). These relations
assume that two answer types encode the same event, and models that
fail the answer-type group of REP tend to fail them too
(\S\ref{sec:taxonomy}); but models that keep the answer-type laws
fail them as well: JevAny-Qwen-4B (direct token) keeps every
answer-type test and scores 38.5 on marginalisation. Three of these
relations also assume that the answer lies in the menu; reading a
Choice as conditioned on its menu while a Noul is not explains few of
their failures (Appendix~\ref{app:menu}). No relation is kept by every
model.

In LOG, the monotonicity of thresholds and quantifiers holds more
often (group mean 79) than the bounds of conjunctions and disjunctions
(73), and the spread is large: CLM keeps monotonicity in 21\% of the
tests, Open-Jev-9B in 96\%. Some bounds cannot be violated by the
answers a test receives: the $n$-ary conjunction's lower bound
$\sum_ip_{A_i}-(n-1)$ is at most 0 in \vacAndLower\% of its tests and the
disjunction's upper bound $\sum_ip_{A_i}$ at least 1 in \vacOrUpper\% (for
the counting bounds, \vacCountLower\% and \vacCountUpper\%), so the high
scores of these two relations say little about logical coherence. In
CHO, three models keep independence of irrelevant alternatives in
every test (Von, CLM and Open-Jev-9B), while Verdict keeps it in a
third of the tests.

\subsection{Coherence and accuracy}
\label{sec:accuracy}

\begin{figure}[t]
\centering
\includegraphics[width=0.5\linewidth]{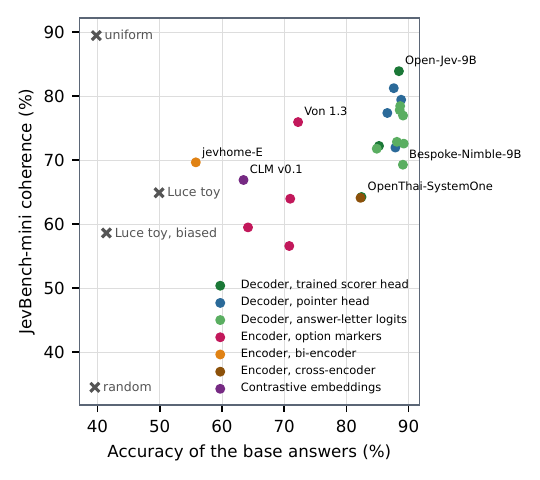}
\caption{Coherence on JevBench-mini plotted against base-answer accuracy.
Crosses indicate reference models. The uniform model has high
coherence at chance accuracy. Among decoders, coherence varies more
than accuracy.}
\label{fig:accuracy}
\end{figure}

Coherence is necessary for a trustworthy decision model, not
sufficient (\S\ref{sec:jevbench_definition}). The reference models
make this concrete (Table~\ref{tab:leaderboard},
Figure~\ref{fig:accuracy}).

\paragraph{Why a model that ignores its input scores \baseUniform.}
The uniform model assigns equal probability to each possible answer:
one half for each Noul answer and one quarter for each option or level
of a four-option Choice or four-level Score. Its output depends only on
the number of possible answers. It satisfies laws that preserve this
uniform allocation and can fail laws that require reallocating
probability according to the events each answer represents.
\begin{itemize}
    \item \emph{Laws that ask an answer to stay the same, to keep an
    order, or to condition on a menu.} Every REP law compares two
    descriptions of the same events, every BAT law the same question
    with different companions, every LOG law asks probabilities to
    respect an order, and every CHO law asks the distribution over the
    remaining options to be the renormalised original. A uniform answer
    keeps all of them: two descriptions of a four-option Choice both
    get one quarter per option, all propositions of a chain get one
    half (an order holds with equality), and a uniform distribution
    renormalised to any sub-menu is the uniform distribution of that
    sub-menu. REP, BAT, LOG and CHO all score 100.
    \item \emph{Laws that ask probability to be split according to what
    the answers mean.} Most MEA laws compare different events of one
    measure: merging two options must add their probabilities,
    splitting one must divide it, a question about a single option must
    match that option's share. These laws fix how much the answer must
    move when the question changes, and the right amount depends on
    which events the options denote. The uniform model moves by an
    amount set only by the number of options, and fails every such law
    whose answer differs from a uniform split (Table~\ref{tab:uniform}).
\end{itemize}

\begin{table}[t]
\centering
\caption{The nine relations the uniform model fails, on a four-option
Choice or a four-level Score, the typical questions of the benchmark.
Each law fixes a value from the source answer; the uniform model
answers from the number of options alone. $\delta$: violation measure
(\S\ref{sec:measure}), against the tolerance $0.05$.}
\label{tab:uniform}
\small
\begin{tabular}{@{}l>{\raggedright\arraybackslash}p{5.5cm}>{\raggedright\arraybackslash}p{4.6cm}c@{}}
\toprule
Relation & What the law requires & Uniform answer & $\delta$\\
\midrule
\texttt{option\_merge} & merged option $a\cup b$: $\tfrac14+\tfrac14=\tfrac12$, others $\tfrac14$ & three options, $\tfrac13$ each & 0.17\\
\texttt{option\_split} & the two parts of $o$ sum to $P_q(o)=\tfrac14$ & five options, parts sum to $\tfrac25$ & 0.15\\
\texttt{option\_clone} & $o$ and its copy sum to $\tfrac14$ & five options, sum $\tfrac25$ & 0.15\\
\texttt{similarity\_effect} & $b$ and its near-duplicate sum to $\tfrac14$ & five options, sum $\tfrac25$ & 0.15\\
\texttt{extreme\_level} & new top level and old top sum to $\tfrac14$ & five levels, sum $\tfrac25$ & 0.15\\
\texttt{choice\_indicator} & ``is it $o$?'' $=P_q(o)=\tfrac14$ & $\tfrac12$ & 0.25\\
\texttt{choice\_complement} & ``not $o$?'' $=1-P_q(o)=\tfrac34$ & $\tfrac12$ & 0.25\\
\texttt{score\_complement} & ``not level $k$?'' $=1-P_q(k)=\tfrac34$ & $\tfrac12$ & 0.25\\
\texttt{score\_cumulative} & ``level $k$ or higher?'' $=\tfrac34,\tfrac12,\tfrac14$ & $\tfrac12$ for every $k$ & 0.25\\
\bottomrule
\end{tabular}
\end{table}

The other eight MEA relations hold for a uniform answer on the tests
of JevBench-mini: a Noul and its negation, templated, natural or of a
conjunction, give $\tfrac12+\tfrac12=1$; inclusion--exclusion gives
$\tfrac12+\tfrac12=\tfrac12+\tfrac12$; a uniform four-way Choice over the
answers to two Nouls has marginals $\tfrac12$, equal to the Nouls;
replacing an option by a catch-all keeps four options at one quarter
each; and the coarsenings of these tests merge four outcomes into two
blocks of two, so that the merged answer, one half per block, is
again uniform. MEA is therefore $8/17=47.1$, and the overall score,
the mean of the five dimensions, is $(4\times100+47.1)/5=89.4$.

The high score reflects the laws that uniform answers satisfy.
Each dimension tests a different operation on the model's outputs, so
we give the five dimensions equal weight. We report accuracy and the
Brier score separately from coherence to show how informative those
outputs are.

\paragraph{Flattened answers.}
Increasing the temperature softens a model's outputs. Renormalising
$P(y)^{1/T}$ preserves the most probable answer and therefore
accuracy. As $T$ grows, the probabilities approach the uniform
reference and the coherence score approaches \baseUniform. A proper
score can detect this change in the distribution. Tempering is not in itself a fault: a model
whose temperature is fitted for calibration \cite{guo2017calibration}
lowers its Brier score, while flattening beyond that point raises it,
so coherence and the Brier score read together tell the two apart. The
Brier score of the base answers (Table~\ref{tab:leaderboard}) is
\brierUniform{} for the uniform model and \brierMin--\brierMax{} for the
evaluated models; the log loss ranks the models almost identically
(Spearman \logLossBrierSpearman{} with the Brier score) and is
\logLossUniform{} for the uniform model. Across models a lower Brier
score goes with higher coherence (Spearman \brierCohSpearman) and higher
accuracy (\brierAccSpearman).

\paragraph{The accuracy of the reference models.}
Accuracy is computed from the base answers alone
(\S\ref{sec:models}). For the uniform model every answer of a question
ties for the highest probability, so each question earns $1/k$ for its
$k$ answers: one half for each of the 1{,}152 Nouls with a known gold
answer, one quarter for each of the 548 four-option Choices and 222
four-level Scores, and one fifth for each of the 16 five-level Scores.
The mean over the \accN{} questions is \baseAccUniform\%, the accuracy
of guessing. The random model, whose answers to different questions
are unrelated, scores \baseRandom{} at \baseAccRandom\% accuracy; only
BAT, which compares identical questions, holds in full. The two toy
scorers fall between (\baseFakeCoherent{} and \baseFakeBiased{}, at
\baseAccFakeCoherent\% and \baseAccFakeBiased\%).

\paragraph{Coherence and accuracy of the models.}
Across the \nModels{} models, accuracy ranges from \accMin\% to
\accMax\%, and its rank correlation with coherence is
\accCohSpearman. The correlation comes mostly from the families: the
option-marker encoders, the bi-encoder and CLM are the least accurate
(55.8--72.2\%) and mostly less coherent. Within the \nDecoder{} decoders, accuracy lies in a narrow
band (\accDecoderMin--\accDecoderMax\%) while coherence spans
\cohDecoderMin--\cohDecoderMax. OneJev-4B, Jebadiah 9B and
Bespoke-Nimble-9B are the three most accurate models (89.1--89.2\%)
and are ranked tenth, seventh and fifteenth on coherence. Some models
reverse the order of the two measures: jevhome-E, a 48M bi-encoder,
is more coherent than the 1B cross-encoder jevhome-ettin-1b distilled
from the same teacher (69.6 against 64.1) and far less accurate (55.8\%
against 82.3\%). The two measures thus answer different questions, and
a model should be read on both: coherence without accuracy can be
achieved by ignoring the input, and accuracy without coherence leaves
the answers inconsistent once they are combined.

\subsection{Batch independence}
\label{sec:batch}

\begin{table}[t]
\centering
\caption{Batch-independence relations (scores in \%) for the two
models that fail them, and the lowest score of the other models.}
\label{tab:bat}
\small
\adjustbox{max width=\linewidth}{\begin{tabular}{@{}lccccc@{}}
\toprule
Model & \texttt{batch solo} & \texttt{batch order} & \texttt{duplicate question} & \texttt{batch content} & \texttt{batch size sweep}\\
\midrule
Bespoke-Nimble-9B & 87.5 & 4.2 & 100.0 & 91.7 & 58.3\\
OpenThai-SystemOne & 83.3 & 12.5 & 66.7 & 91.7 & 100.0\\
Other 19 models (lowest) & 100.0 & 100.0 & 100.0 & 100.0 & 100.0\\
\bottomrule
\end{tabular}
}
\end{table}

Nineteen models pass every batch-independence test at the reported
tolerance. Processing questions independently makes answers invariant
to the other questions in a request. The two models that fail these
tests process the batch differently (Table~\ref{tab:bat}).

OpenThai-SystemOne concatenates the state and all questions into one
sequence and reads each answer at the answer token of its question.
The backbone is causal, so each answer depends on the questions
\emph{before} it. Reversing the batch changes most answers
(\texttt{batch\_order} holds in 13\% of tests), and a question asked
twice in one request is often answered differently the second time
(\texttt{duplicate\_question}, 67\%), because its second copy sees the
first. Adding or removing questions at the end leaves the earlier
answers unchanged (\texttt{batch\_size\_sweep}, 100\%).

Bespoke-Nimble-9B shows the whole schema of the request, every
question with its options, in the prompt of each question, and scores
the requested one. Each answer therefore depends on \emph{all} the
other questions, their order and their number: \texttt{batch\_order}
holds in 4\% of tests and \texttt{batch\_size\_sweep} in 58\%. A
question asked twice is answered the same both times, since the two
prompts differ only in the name of the requested field. We checked
that our adapter builds the same schema as the developers' own request
compiler, so this is a property of the model, not of the serving.

Reading all questions in one pass can reduce cost. Including other
questions can help a model interpret the current one. In both designs,
however, answers depend on how the client batches questions, and a
decision system rarely controls that choice. For Bespoke-Nimble-9B,
reordering changed an answer by more than 0.9 in total variation in
the worst test.

\subsection{Comparisons between related releases}
\label{sec:controlled}

Related releases provide comparisons of readouts, backbones, task
fine-tuning and scale. Differences in training, size or serving
configurations can also contribute to the observed score differences.
\begin{itemize}
    \item \emph{Readout and backbone, same data.} The three JevAny
    models were trained on the same 1.77M records. With a pointer
    readout, the Gemma backbone (81.2) and the Qwen backbone (79.4)
    are not separated; on the Qwen backbone, the pointer readout and
    the direct-token readout (77.8) are not separated overall; their
    largest difference is in CHO (84.7 against 76.4).
    \item \emph{Task fine-tuning.} Laya typed-decisions is Laya
    fine-tuned on four workflows. Coherence rises from 56.5 to 63.9 at
    similar accuracy (70.8\% and 71.0\%).
    \item \emph{Scale.} decider-4b is more coherent than decider-2b
    (78.4 against 71.7) and more accurate (88.6\% against 84.9\%).
    Across all models, the rank correlation between parameter count
    and coherence is \sizeSpearman.
    \item \emph{Bi-encoder or cross-encoder.} The jevhome pair is
    discussed in \S\ref{sec:accuracy}: in this pair, the bi-encoder
    has the higher coherence and the cross-encoder the higher accuracy;
    the two also differ in size (48M and 1B).
\end{itemize}

\subsection{JevBench-mini against JevBench-240}
\label{sec:mini}

\begin{figure}[t]
\centering
\begin{minipage}{0.42\linewidth}
\centering
\includegraphics[width=\linewidth]{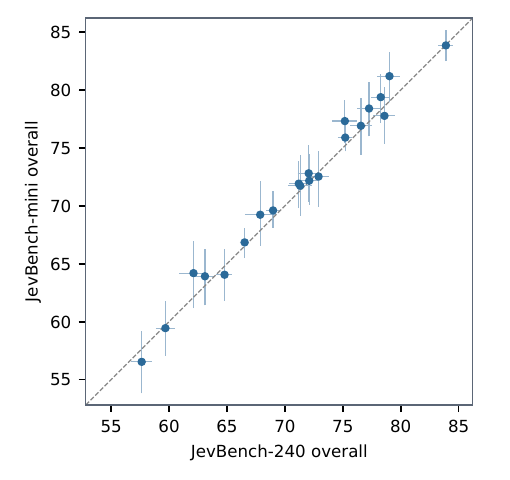}
\end{minipage}\hfill
\begin{minipage}{0.55\linewidth}
\centering
\small
\adjustbox{max width=\linewidth}{\begin{tabular}{@{}lcccc@{}}
\toprule
Score & Spearman & Pearson & mean $|\Delta|$ & max $|\Delta|$\\
\midrule
Overall & 0.983 & 0.992 & 0.86 & 2.17\\
REP & 0.981 & 0.982 & 2.00 & 5.28\\
BAT & 0.837 & 0.998 & 0.48 & 7.17\\
MEA & 0.962 & 0.994 & 1.17 & 3.38\\
LOG & 0.968 & 0.980 & 1.61 & 4.12\\
CHO & 0.974 & 0.981 & 2.89 & 8.33\\
\bottomrule
\end{tabular}
}
\end{minipage}
\caption{Overall coherence of the \agreeN{} models run on both suites,
with 95\% intervals (left), and agreement of the overall and dimension
scores (right; $\Delta$: JevBench-mini minus JevBench-240, in points).}
\label{fig:mini}
\end{figure}

JevBench-mini sends a ninth of the requests of JevBench-240. On the
\agreeN{} models run on both (Figure~\ref{fig:mini}), the overall
scores agree closely: Pearson \agreePearson, Spearman \agreeSpearman,
Kendall \agreeKendall. JevBench-mini scores are higher by \agreeMeanDiff{} points on
average. The largest absolute difference is \agreeMaxAbsDiff{}
points, and the largest rank shift is \agreeMaxRankShift{} places. Of the
\agreePairs{} pairs of models, \agreePairsSame{} are ordered the same
way, and \agreeSepSame{} of the \agreeSep{} pairs that JevBench-240
separates (95\% intervals apart) are ordered the same way by
JevBench-mini; the exception, jevhome-ettin-1b and OpenThai-SystemOne,
is 0.1 points apart on JevBench-mini. The
dimension scores agree as well, except that the rank correlation of BAT
is lower (\agreeBATSpearman) only because most models tie at 100. The smaller suite produces intervals about \ciRatio{} times as
wide (\ciWidthMini{} compared with \ciWidthFull{} points on average).
JevBench-240 can distinguish models that are a few points apart when
their JevBench-mini intervals overlap.

\paragraph{A disjoint sample.}
Every JevBench-mini test is a JevBench-240 test, so the agreement of
the two suites is not an independent check. Scored on the
\splitTests{} JevBench-240 tests outside JevBench-mini, the \splitN{}
models rank as on JevBench-mini (Spearman \splitSpearman, Pearson
\splitPearson).

\paragraph{Paired comparisons.}
Table~\ref{tab:leaderboard} separates two models when their intervals
do not overlap, which is conservative. All models answer the same
cases, so a paired bootstrap, which resamples the cases once for all
models and takes the interval of each difference, is the direct test:
on JevBench-mini it separates \pairsPaired{} of the \pairsTotal{} pairs,
against \pairsOverlap{} by overlap. \topModel{} leads JevAny-Gemma-4B
by \pairTopDiff{} points (95\% interval \pairTopLo--\pairTopHi) and is
separated from all \pairTopSeparated{} other models.

\paragraph{Weights.}
The overall score weights the five dimensions equally, so one CHO
relation carries $1/15$ of it and one MEA relation $1/85$, and BAT,
saturated for \nBATperfect{} models, carries a fifth. Weighting every
relation equally instead gives a ranking with Spearman
\wEqualSpearman{} to the reported one; leaving out any one dimension
gives \wDropMin--\wDropMax. The first model is \wTopUnchanged{} in every
case.

\paragraph{Domains.}

We do not compare models by domain. JevBench-mini tests each relation
on two cases of a domain, fewer than the five a relation needs to be
scored there. JevBench-240 has 20 cases per domain, but on the
\domModels{} models run on it a domain score differs from the model's
overall score by only \domDeviation{} points on average, while its 95\%
interval (resampling the domain's cases) spans $\pm$\domHalfWidth{}
points; the overall score falls outside that interval for
\domOutside\% of the (model, domain) pairs. With 20 cases per domain, these intervals are too wide to distinguish
a model's domain scores.

\subsection{Sensitivity to the tolerance}
\label{sec:tolerance}

\begin{figure}[t]
\centering
\includegraphics[width=0.56\linewidth]{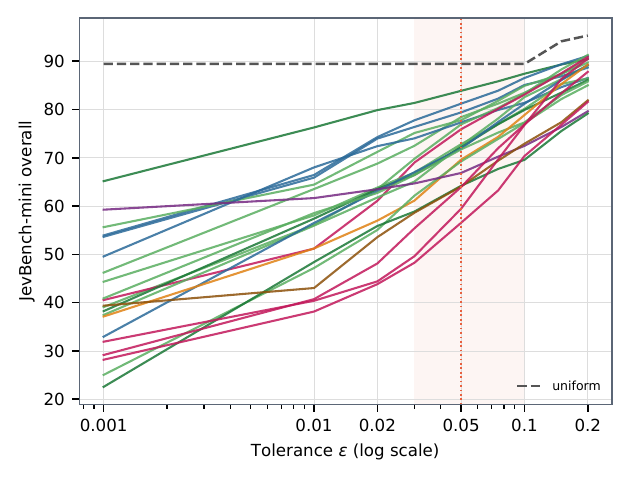}
\caption{Overall coherence on JevBench-mini as the tolerance varies, one
line per model, coloured by family; dashed, the uniform reference. Scores are nondecreasing in the tolerance (Proposition~\ref{prop:relaxation}); the
shaded band, $0.03\le\varepsilon\le0.1$, is where the ranking barely
changes. Dotted, the reported tolerance.}
\label{fig:tolerance}
\end{figure}

A test passes when its violation is at most $\varepsilon=0.05$ in
probability. Since every test's violation is stored, the scores at any
other tolerance follow from the same reports without a further request
(Figure~\ref{fig:tolerance}, Table~\ref{tab:tolerance} in
Appendix~\ref{app:tolerance}). Between $\varepsilon=0.03$ and $0.1$ the
ranking of the \nModels{} models barely changes: its rank correlation
with the ranking at $0.05$ is at least \tolBandSpearmanMin, no model
moves by more than \tolBandShiftMax{} ranks, and the first model stays
first. Averaging each model's score over all the tolerances we tried,
from $0.001$ to $0.2$, gives a ranking with rank correlation
\tolAverageSpearman{} to the reported one, so the ranking does not rest
on a lucky choice of $\varepsilon$.

At the extremes, the tolerance changes which violations drive the
ranking. At $0.001$ (mean
\tolStrictMean), a test passes only if the law holds to the third
decimal, which rewards architectures that keep some laws exactly by
construction: CLM, which scores each option independently, rises by
\tolClmStrictGain{} ranks. At $0.2$ (mean \tolLooseMean), violations up
to a fifth of the probability pass; the scores crowd together and
Verdict, two thirds of whose violations lie between 0.05 and 0.2, rises by
\tolVerdictLooseGain{} ranks (rank correlation with the reported
ranking \tolLooseSpearman). The uniform reference scores
\baseUniform{} up to $\varepsilon=0.1$, since the laws it breaks miss
by $0.15$ to $0.25$ (Table~\ref{tab:uniform}), and stays above every
model at every tolerance (\tolUniformLoose{} at $0.2$).

\subsection{Cost}
\label{sec:cost}

A JevBench-mini run sends \nMiniRequests{} requests. On one replica,
the fastest model answers in \fastestSecPerReq{} seconds per request,
under a minute for the whole suite, and the slowest in
\slowestSecPerReq{} seconds, about two hours
(Table~\ref{tab:runtime} in Appendix~\ref{app:runtime}). The cost is
set by how a server answers a request: one forward pass per question
or per option, a sequential CPU runtime, or several passes per
question. JevBench-240 takes about eight times as long. The runs of
Table~\ref{tab:runtime}, every model on both suites, take about
\gpuHours{} GPU-hours on one A100 40GB in total (and \cpuHours{}
CPU-hours for the two jevhome models), and generating the data, the
rewrite bank included, took about 3 GPU-hours. Requests are cached, so
changing the scoring requires no new requests.


\section{Discussion}
\label{sec:discussion}

\paragraph{What a coherence score means.}
A relation score is the fraction of tests that satisfy the law within
0.05 in probability. Uniform answers can score well because they
satisfy laws that do not require a particular allocation of
probability. We therefore report coherence alongside accuracy and the
Brier score (Section~\ref{sec:accuracy}). Among the decoder models,
accuracy falls within a relatively narrow range
(\accDecoderMin--\accDecoderMax\%), while coherence spans about
twenty points.

\paragraph{Additivity across questions.}
Probability-measure coherence is the lowest-scoring dimension for
every evaluated model.
These are the laws a downstream system relies on most directly: it
negates conditions, merges options and compares a yes/no question with
the Choice it summarises. Our results suggest that current training
objectives, which fit each question's distribution to a label, do not
by themselves make the answers to related questions add up. Training
on transformed questions with consistency targets, as has been
proposed for question answering \cite{li2019logic}, is a natural
direction; the frozen relations of JevBench could serve as such
targets, provided the benchmark's own cases are kept out of training.

\paragraph{Batching is part of the model.}
Two of the models make each answer depend on the other questions of a
request. The interface allows a client to send any batch, so a client
cannot know that its batching changes the answers unless the model
documents it. The pattern of failed batch relations distinguishes the two observed
failure modes: dependence on earlier questions and dependence on all
questions in the batch.

\section{Limitations}
\label{sec:limitations}

\begin{itemize}
    \item \emph{Synthetic, English data.} The cases are generated by one
    writer model and checked by one verifier, in English, in twelve
    business domains. The cases are balanced by construction and the verifier checks their
    facts. The dataset represents synthetic scenarios, so results may
    differ on real traffic. Models reviewed the rewrite bank, and we
    also inspected a sample. The reliability of the gold answers
    depends on the verifier and the answer rules. The spot check found
    ambiguity, particularly in Scores, that affects the accuracy
    comparison (Section~\ref{sec:cases}). A law
    holds whatever the answer to a vague question is; coherence is at
    risk where a transformation may change a question's meaning, as a
    rewrite can (\spotRewriteValid{} of \spotRewriteN{} sampled rewrites
    were valid).
    \item \emph{One fixed tolerance.} All laws are judged at 0.05 in
    probability. The ranking is stable from 0.03 to 0.1 and changes at
    the extremes, where the scores reward exact symmetries or forgive
    large violations (\S\ref{sec:tolerance}); the package rescores saved
    results at any tolerance, and reports a graded score that measures
    how far beyond the tolerance violations go.
    \item \emph{One test per relation and case.} A relation with
    several variants is tested on one of them per case, in turn. Its
    score estimates the share of (case, variant) pairs at which the law
    holds, which is never lower than the share of cases at which every
    variant holds (\S\ref{sec:measure}).
    \item \emph{The reported subset.} JevBench-mini agrees with
    JevBench-240 on the ranking of all \nFull{} models, but its
    intervals are about \ciRatio{} times as wide. Neither suite
    separates a model's domains (\S\ref{sec:mini}).
    \item \emph{Model coverage.} We evaluated openly released models built
    for the Jev interface. General-purpose instruction-tuned language
    models are not among them; any of them can be tested through the same
    interface, for example by reading the probabilities of the answer
    letters, and doing so is the natural next step.
    \item \emph{Serving.} We served each model with its developers'
    code at a pinned revision. Two models needed a third-party server
    or our adapter, and every model's results reflect its released
    serving configuration (precision, prompt format, temperatures), not
    the best configuration it could have.
    \item \emph{Repeatable answers.} No self-hosted model varied across
    three repetitions of the same request under the tested serving
    configurations, so a test compares one answer with one answer. A
    model that samples would add its own noise to every test; the
    package can measure that noise by repeating requests, but the
    benchmark judges such a model at the same tolerance.
    \item \emph{A public test set.} The cases and tests are public. A
    model trained on them would no longer be measured fairly; the data
    generator is part of the package, so new cases can be generated
    for a private evaluation.
\end{itemize}

\section{Ethical considerations}
\label{sec:ethics}

The cases are synthetic and describe fictional people and organisations;
they contain no personal data and are not designed to represent any
demographic group. We scanned all 240 states for profanity,
slurs and other offensive terms and found none; the content-moderation
cases describe policy violations without quoting abusive text. The
evaluated models were used under their licences, for research. Coherence measures whether a model's
answers are consistent with one another, not whether they are correct
or fair, and a high score should not be read as fitness for a decision
task: a model that ignores its input scores higher than every model we
evaluated (\S\ref{sec:accuracy}). Because the test set is public, scores
of models trained on it would not be comparable (\S\ref{sec:limitations}).

\section{Conclusion}
\label{sec:conclusion}

JevBench evaluates typed probabilistic decision models using 50 laws
of probability and choice and the same fixed requests for every
model. The coherence tests require no gold labels. Among \nModels{}
openly released models, most satisfy batch independence, while
additivity violations are common, especially for negations. A uniform
reference scores above all evaluated models on coherence while
achieving chance accuracy. Reporting both measures distinguishes
consistency across questions from accuracy on individual questions.
The benchmark, data generator and results are publicly available.

\section*{Acknowledgements}
The research use of models is described in Section~\ref{sec:benchmark}.
gpt-oss-20b generated the data, Llama-3.1-8B-Instruct verified it,
Claude reviewed the rewrites and GPT-5.6 Sol performed the spot check.
Claude Code also helped write and test the package, analysis scripts
and project website, and helped draft and edit the paper. ChatGPT
reviewed drafts. The author designed the
benchmark, checked all code, numbers and claims, and takes full
responsibility for the content.

\bibliographystyle{plain}
\bibliography{refs}

\appendix

\section{Candidate laws excluded from the benchmark}
\label{app:excluded}

We group the excluded candidates by reason: implication by another
law on the same outputs (Section~\ref{app:implied}), overlap with
another relation on some tests (Section~\ref{app:coincide}), or
failure to meet Definition~\ref{def:relation}
(Section~\ref{app:invalid}). Notation follows Section~\ref{sec:method}.
Here $P_q$ is the source output, $P_{q'}$ is the follow-up output, $p$
is a Noul's probability of yes, $C$ is the original option set and
$P|_K$ is a distribution renormalised to $K$.

For each implied law we give the derivation at $\varepsilon=0$ and a
bound on its violation measure $\delta_{\mathrm w}$ in terms of the
measure $\delta_{\mathrm s}$ of the relation that implies it. When
$\delta_{\mathrm w}\le\delta_{\mathrm s}$, every test that passes the
stronger relation at tolerance $\varepsilon$ passes the weaker law at
the same tolerance; when the bound carries a factor, the implication
holds exactly but may fail at a positive tolerance. We repeatedly use
that a pointwise difference never exceeds the total variation,
$|P(y)-Q(y)|\le\TV(P,Q)$, and that for $f:\mathcal{Y}\to[0,1]$,
$|\mathbb{E}_Pf-\mathbb{E}_Qf|\le\TV(P,Q)$.

\subsection{Laws implied by other relations}
\label{app:implied}

\paragraph{Regularity after an addition.}
When a menu grows by an alternative $z$, no original alternative should
gain probability: $P_{q'}(a)\le P_q(a)$ for every $a\in C$. Independence
of irrelevant alternatives, $P_{q'}|_C=P_q$, gives
\begin{equation}
    P_{q'}(a)=\bigl(1-P_{q'}(z)\bigr)P_{q'}|_C(a)=\bigl(1-P_{q'}(z)\bigr)P_q(a)\le P_q(a),
\end{equation}
and in general
$P_{q'}(a)-P_q(a)\le P_{q'}|_C(a)-P_q(a)\le\TV(P_{q'}|_C,P_q)$, so
$\delta_{\mathrm w}\le\delta_{\mathrm s}$ for \texttt{irrelevant\_option}.
When the addition refines an option $b$, as a near-duplicate $b'$
(\texttt{similarity\_effect}) or a level beyond the top of a Score
(\texttt{extreme\_level}), the partition law keeps every other outcome
and gives $P_{q'}(b)+P_{q'}(b')=P_q(b)$, hence $P_{q'}(b)\le P_q(b)$; each
coordinate of $\phi_{\#}P_{q'}-P_q$ is at most $\delta_{\mathrm s}$, so
again $\delta_{\mathrm w}\le\delta_{\mathrm s}$.

\paragraph{Regularity after a removal.}
When an option $o$ is removed, no remaining alternative should lose
probability: $P_q(a)\le P_{q'_o}(a)$. Independence,
$P_{q'_o}=P_q|_{C\setminus\{o\}}$, gives
$P_{q'_o}(a)=P_q(a)/(1-P_q(o))\ge P_q(a)$; since
$P_q(a)\le P_q|_{C\setminus\{o\}}(a)$, also
$P_q(a)-P_{q'_o}(a)\le P_q|_{C\setminus\{o\}}(a)-P_{q'_o}(a)\le\delta_{\mathrm s}$
for \texttt{remove\_option\_iia}.

\paragraph{Compromise effect.}
A level beyond the top of a Score should not draw probability towards
the lower levels: $P_{q'}(k)\le P_q(k)$ for every level $k$ below the
top. Under \texttt{extreme\_level} every lower level keeps its
probability, and $P_{q'}(k)-P_q(k)\le\TV(\phi_{\#}P_{q'},P_q)=\delta_{\mathrm s}$.

\paragraph{Exclusivity of option Nouls.}
The yes probabilities of the Nouls ``is it $o$?'', one for each
option, should sum to one. With
$p_{q'_o}=P_q(o)$ for every option (\texttt{choice\_indicator}),
$\sum_op_{q'_o}=\sum_oP_q(o)=1$. In general
\begin{equation}
    \Bigl|\sum_op_{q'_o}-1\Bigr|
    =\Bigl|\sum_o\bigl(p_{q'_o}-P_q(o)\bigr)\Bigr|
    \le k\,\delta_{\mathrm s}
\end{equation}
for $k$ options: the implication is exact, but errors within tolerance
can add up.

\paragraph{Monotonicity of cumulative Nouls.}
The yes probabilities of the Nouls ``level $k$ or higher'' should
be nonincreasing in $k$. Writing
$\tau_k=\sum_{j\ge k}P_q(j)$, which decreases in $k$,
\texttt{score\_cumulative} requires $p_{q'_k}=\tau_k$, and for $k<k'$
\begin{equation}
    p_{q'_{k'}}-p_{q'_k}
    =\bigl(p_{q'_{k'}}-\tau_{k'}\bigr)-\bigl(p_{q'_k}-\tau_k\bigr)+\bigl(\tau_{k'}-\tau_k\bigr)
    \le2\,\delta_{\mathrm s}.
\end{equation}

\paragraph{Three of the four binary Fr\'echet bounds.}
Write $g=p_{A\land B}+p_{A\lor B}-p_A-p_B$ for the inclusion--exclusion
gap (\texttt{inclusion\_exclusion}). Then
\begin{align}
    \text{lower bound of } A\land B:\quad
    & p_A+p_B-1-p_{A\land B}=-g+\bigl(p_{A\lor B}-1\bigr)\le|g|,\\
    \text{upper bound of } A\lor B:\quad
    & p_{A\lor B}-p_A-p_B=g-p_{A\land B}\le|g|,\\
    \text{lower bound of } A\lor B:\quad
    & \max\{p_A,p_B\}-p_{A\lor B}=\bigl(p_{A\land B}-\min\{p_A,p_B\}\bigr)-g.
\end{align}
The first two follow from inclusion--exclusion alone, since
$p_{A\lor B}\le1$ and $p_{A\land B}\ge0$. The third follows from
inclusion--exclusion together with the upper bound of $A\land B$
(\texttt{frechet\_and\_upper}), with
$\delta_{\mathrm w}\le\delta_{\texttt{frechet\_and\_upper}}+|g|$.

\paragraph{Fr\'echet bounds of the pair Choice.}
The four-way Choice over the answers to $A$ and $B$ gives a joint cell
$P_{q'}(11)$, which should satisfy
$p_A+p_B-1\le P_{q'}(11)\le\min\{p_A,p_B\}$. Once its marginals equal
$p_A$ and $p_B$ (\texttt{pair\_choice\_marginals}), both bounds hold
for any distribution over the four cells:
$P_{q'}(11)\le P_{q'}(1\cdot)$ and
$P_{q'}(11)=P_{q'}(1\cdot)+P_{q'}(\cdot1)-1+P_{q'}(00)$.

\paragraph{Restrictions of a relation's own law.}
Three further checks restate part of a relation's law. For a split
option, the parts should sum to the whole,
$P_{q'}(o\land X)+P_{q'}(o\land\neg X)=P_q(o)$; this is one coordinate of
\texttt{option\_split}, $(\phi_{\#}P_{q'})(o)=P_q(o)$. Proportionality of
the renormalised option Nouls to the Choice follows from
\texttt{choice\_indicator}. For a near-duplicate $b'$ of $b$, the
requirement that every other option keep its probability is equivalent
to \texttt{similarity\_effect}, since the mass of $b$ and $b'$ together
is then fixed by the others.

\subsection{Coinciding tests}
\label{app:coincide}

Two relations would share tests with another relation if their index
were not restricted. A batch of size one is the question asked alone,
so \texttt{batch\_size\_sweep} starts at size two and leaves that test
to \texttt{batch\_solo}. With three options, removing one leaves a pair,
so \texttt{pairwise\_iia} applies from four options on and leaves
three-option menus to \texttt{remove\_option\_iia}.

\subsection{Candidate laws outside the definition}
\label{app:invalid}

\paragraph{Laws that need the correct answer.}
Three natural checks refer to the correct answer and so violate (P2):
that the answers to ``more than $k$?'' follow the step
$\mathbb{I}[k<\text{true count}]$; that an off-topic option receives no
probability, which presumes it is wrong; and that a ``none of the
above'' option added to the full menu receives none. The last is not a
law of the model's outputs either: a Choice forces an answer among its
options, so putting all mass on them does not assert that one of them
holds.

\paragraph{Transformations chosen from the model's answers.}
Removing the option the model finds least probable, or replacing the
one it finds most probable by a catch-all, tests the same laws as
\texttt{remove\_option\_iia} and \texttt{option\_closure}. Because the
removed option would depend on the model, different models would
answer different questions, contrary to (P4). Removing or replacing
every option in turn tests these laws on a fixed test set.

\paragraph{Hypothetical conditioning.}
The chain rule $p_{A\land B}=p_A\,p_{B\mid A}$ needs the answer to a
conditional question, ``supposing $A$, is $B$ the case?''. When the
state already settles $A$, such a supposition can contradict the state,
and the conditional question has no answer that the law constrains.

\paragraph{Independence for near-duplicates.}
Independence of irrelevant alternatives is not a coherence law for an
alternative that nearly duplicates an existing one: it would have the
duplicate take share from every option in proportion, where the
duplicate should take it from its original. This is the classical
objection to the choice axiom for similar alternatives
\cite{tversky1972}. A near-duplicate refines its original, and JevBench
tests it with the partition law (\texttt{similarity\_effect}).

\paragraph{Decoys.}
A decoy is a weaker or partial version of a target option $t$. It
overlaps $t$: an answer that fits the decoy also fits $t$, so the menu
no longer lists disjoint events. Neither the conditioning law of CHO
nor a partition law then applies, and the attraction effect has no
coherence law on such a menu. A decoy disjoint from $t$ would be an
ordinary added option, subject to the conditioning law.


\section{Models}
\label{app:models}

Table~\ref{tab:models} lists the evaluated models by family, from their
own documentation and server source. Each was installed at a pinned
revision; the exact installation and serving commands are part of the
benchmark's repository. Table~\ref{tab:families} groups the overall
scores by readout family.

\begin{table}[H]
\centering
\caption{The \nModels{} evaluated models.}
\label{tab:models}
\footnotesize
\setlength{\tabcolsep}{3pt}
\adjustbox{max width=\linewidth}{\begin{tabular}{@{}l>{\raggedright\arraybackslash}p{2.6cm}>{\raggedright\arraybackslash}p{3.0cm}>{\raggedright\arraybackslash}p{5.0cm}r@{}}
\toprule
Model & Developer & Base model & Readout and training & Params\\
\midrule
\multicolumn{5}{@{}l}{\textit{Decoder, trained scorer head}}\\
Open-Jev-9B & Zefan Cai & Qwen3.5-9B & rank-8 LoRA + scalar decision head; NLL + Brier & 9.7B\\
OpenThai-SystemOne & iApp Technology & Qwen3.5-0.8B-Base (Thai continued pre-training) & 256-slot decision head; questions read as one causal sequence & 0.75B\\
RSI-Jev v3.0 2B & Shanghua Gao & Qwen3.5-2B-Base & trained readout/scorer head; SFT then RL, fitted calibration & 2.3B\\
\multicolumn{5}{@{}l}{\textit{Decoder, pointer head}}\\
JevAny-Gemma-4B (pointer) & SimpleJev & Gemma-4-E4B-it & LoRA SFT, pointer readout & E4B\\
JevAny-Qwen-4B (pointer) & SimpleJev & Qwen3.5-4B & LoRA SFT, pointer readout & 4B\\
Kev-4B & Jared Palmer & Qwen3.5-4B-Base & rank-16 LoRA + pointer head & 4.7B\\
TinyJev-4B & AnkitAI & Qwen3-4B-Base & merged LoRA + pointer head (Kev recipe) & 4.0B\\
\multicolumn{5}{@{}l}{\textit{Decoder, answer-letter logits}}\\
Bespoke-Nimble-9B$^{1}$ & Bespoke Labs & Qwen3.5-9B & LoRA scoring one-letter answer tokens; whole schema in every prompt & 9.7B\\
Jebadiah 9B v2 & Frontier Infra & Qwen3.5-9B & merged LoRA, per-type temperatures & 9.65B\\
JevAny-Qwen-4B (direct token) & SimpleJev & Qwen3.5-4B & LoRA SFT, direct-token readout & 4B\\
JevK5 v0.3 & Alibi Serikbay & Qwen3.5-4B & distilled LoRA (merged), option-letter logit readout & 4.2B\\
OneJev-4B & OmniJev & Qwen3.5-4B & full fine-tune (vision tower frozen); option-letter probability & 5.2B\\
decider-2b v10 & Mapika & Qwen3.5-2B-Base & fine-tune + calibration-aware RL; option-letter logits & 1.9B\\
decider-4b v2.1 & Mapika & Qwen3.5-4B-Base & full fine-tune + LoRA stage; option-letter logits, per-type temperatures & 4.2B\\
\multicolumn{5}{@{}l}{\textit{Encoder, option markers}}\\
Laya (english) & Convai Innovations & ModernBERT-large & one [MASK] marker per option & 421M\\
Laya (typed-decisions) & Convai Innovations & ModernBERT-large & as Laya, fine-tuned on four workflows & 421M\\
Verdict 1.4$^{2}$ & Heman10x & ModernBERT-base & GLiClass bi-encoder head, RLCD/Brier calibration & 151M\\
Von 1.3 & wfzyx & ModernBERT-large & option-marker head; listwise CE + Brier, temperature map & 395M\\
\multicolumn{5}{@{}l}{\textit{Encoder, bi-encoder}}\\
jevhome-E & Jibril Frej & granite-embedding-small-r2 & bi-encoder distilled from a Qwen3.5-4B teacher & 48M\\
\multicolumn{5}{@{}l}{\textit{Encoder, cross-encoder}}\\
jevhome-ettin-1b & Jibril Frej & Ettin-encoder-1B & cross-encoder distilled from a Qwen3.5-4B teacher & 1.0B\\
\multicolumn{5}{@{}l}{\textit{Contrastive embeddings}}\\
CLM v0.1 & Contrastive-LM & Qwen3-8B (frozen) & projection heads over frozen embeddings & 8.2B + 20M\\
\bottomrule
\end{tabular}
}
\par\smallskip
\parbox{0.95\linewidth}{\footnotesize$^{1}$\,Bespoke-Nimble-9B: JevBench shim (Nimble's own Modal server does not run locally); same request mapping as Nimble's compiler.py, PyTorch reference runner, current checkpoint. $^{2}$\,Verdict 1.4: OpenJev server (razorback16/openjev), which reimplements Verdict 1.4's prompt and temperatures; Verdict ships no HTTP server.
}
\end{table}

\begin{table}[H]
\centering
\caption{Overall coherence on JevBench-mini by readout family: number of
models, mean and range.}
\label{tab:families}
\small
\begin{tabular}{@{}lrrr@{}}
\toprule
Readout family & Models & Mean & Range\\
\midrule
Decoder, trained scorer head & 3 & 73.4 & 64.2--83.9\\
Decoder, pointer head & 4 & 77.5 & 71.9--81.2\\
Decoder, answer-letter logits & 7 & 74.2 & 69.2--78.4\\
Encoder, option markers & 4 & 63.9 & 56.5--75.9\\
Encoder, bi-encoder & 1 & 69.6 & 69.6--69.6\\
Encoder, cross-encoder & 1 & 64.1 & 64.1--64.1\\
Contrastive embeddings & 1 & 66.8 & 66.8--66.8\\
\bottomrule
\end{tabular}

\end{table}

\section{Scores by relation}
\label{app:relations}

\begin{figure}[H]
\centering
\includegraphics[width=\linewidth,height=0.82\textheight,keepaspectratio]{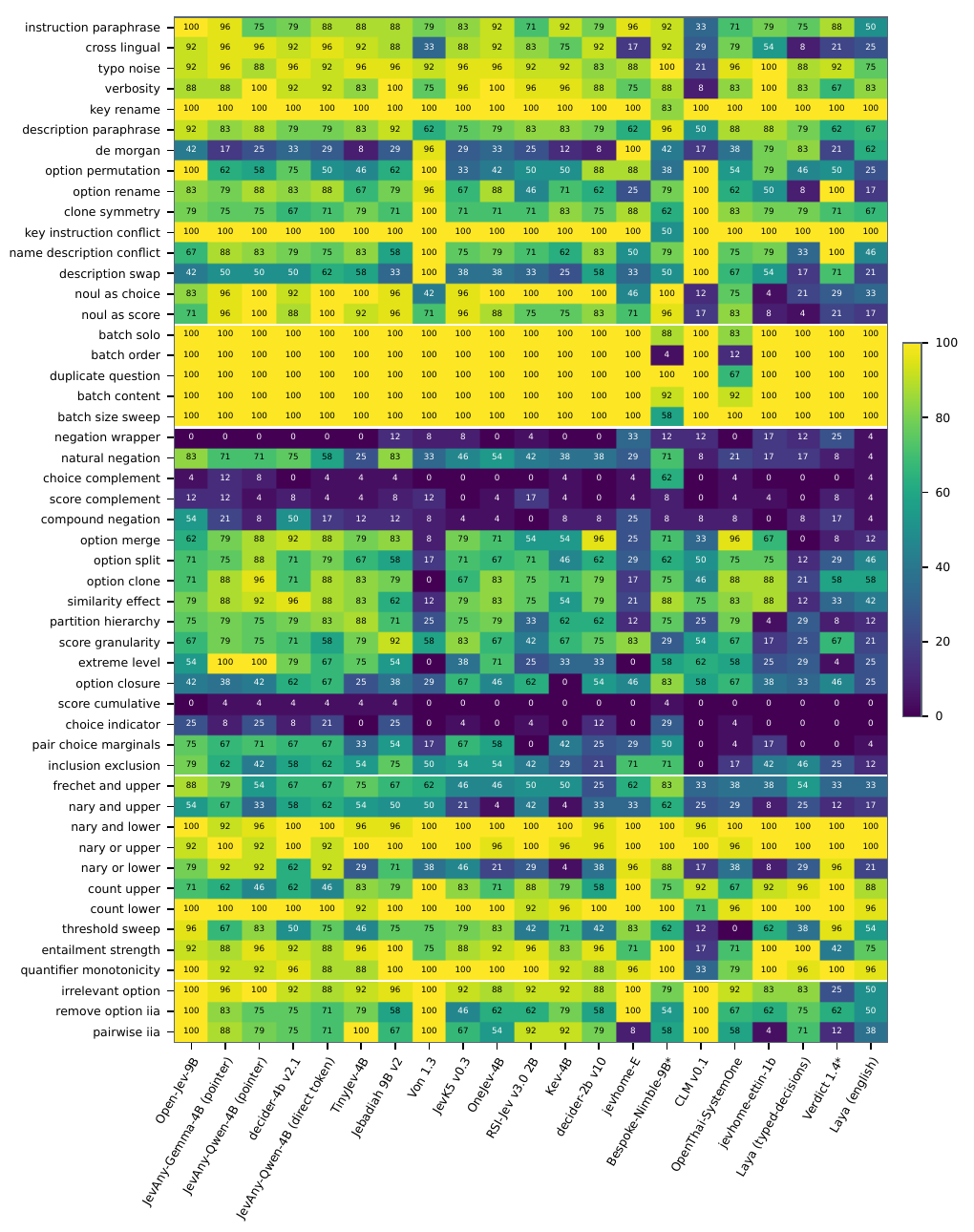}
\caption{Relation scores on JevBench-mini (\%), one column per model,
ordered by overall coherence. White lines separate the dimensions, in
the order REP, BAT, MEA, LOG, CHO. $^\ast$: not served by its
developers' own server.}
\label{fig:relations}
\end{figure}

Figure~\ref{fig:relations} shows each model's scores by relation on
JevBench-mini. Table~\ref{tab:relation_summary} summarises the scores
across models for each relation.

{\small
\begin{longtable}{@{}llrrrr@{}}
\caption{Relation scores on JevBench-mini across the 21 models: mean, lowest and highest, and the number of models at 100.}\label{tab:relation_summary}\\
\toprule
Group & Relation & Mean & Min & Max & \# at 100\\
\midrule
\endfirsthead
\toprule
Group & Relation & Mean & Min & Max & \# at 100\\
\midrule
\endhead
\bottomrule
\endfoot
REP.invariance & \texttt{instruction\_paraphrase} & 80.6 & 33.3 & 100.0 & 1\\
 & \texttt{cross\_lingual} & 68.5 & 8.3 & 95.8 & 0\\
 & \texttt{typo\_noise} & 88.7 & 20.8 & 100.0 & 2\\
 & \texttt{verbosity} & 84.7 & 8.3 & 100.0 & 4\\
 & \texttt{key\_rename} & 99.2 & 83.3 & 100.0 & 20\\
 & \texttt{description\_paraphrase} & 78.6 & 50.0 & 95.8 & 0\\
 & \texttt{de\_morgan} & 39.5 & 8.3 & 100.0 & 1\\
REP.equivariance & \texttt{option\_permutation} & 61.7 & 25.0 & 100.0 & 3\\
 & \texttt{option\_rename} & 68.5 & 8.3 & 100.0 & 2\\
 & \texttt{clone\_symmetry} & 77.0 & 62.5 & 100.0 & 2\\
REP.binding & \texttt{key\_instruction\_conflict} & 97.6 & 50.0 & 100.0 & 20\\
 & \texttt{name\_description\_conflict} & 74.6 & 33.3 & 100.0 & 3\\
 & \texttt{description\_swap} & 50.0 & 16.7 & 100.0 & 2\\
REP.type & \texttt{noul\_as\_choice} & 72.6 & 4.2 & 100.0 & 8\\
 & \texttt{noul\_as\_score} & 68.8 & 4.2 & 100.0 & 2\\
BAT.isolation & \texttt{batch\_solo} & 98.6 & 83.3 & 100.0 & 19\\
 & \texttt{batch\_order} & 91.3 & 4.2 & 100.0 & 19\\
 & \texttt{duplicate\_question} & 98.4 & 66.7 & 100.0 & 20\\
 & \texttt{batch\_content} & 99.2 & 91.7 & 100.0 & 19\\
 & \texttt{batch\_size\_sweep} & 98.0 & 58.3 & 100.0 & 20\\
MEA.complement & \texttt{negation\_wrapper} & 7.1 & 0.0 & 33.3 & 0\\
 & \texttt{natural\_negation} & 42.5 & 4.2 & 83.3 & 0\\
 & \texttt{choice\_complement} & 5.6 & 0.0 & 62.5 & 0\\
 & \texttt{score\_complement} & 6.0 & 0.0 & 16.7 & 0\\
 & \texttt{compound\_negation} & 13.7 & 0.0 & 54.2 & 0\\
MEA.partition & \texttt{option\_merge} & 59.3 & 0.0 & 95.8 & 0\\
 & \texttt{option\_split} & 58.1 & 12.5 & 87.5 & 0\\
 & \texttt{option\_clone} & 66.7 & 0.0 & 95.8 & 0\\
 & \texttt{similarity\_effect} & 67.3 & 12.5 & 95.8 & 0\\
 & \texttt{partition\_hierarchy} & 54.0 & 4.2 & 87.5 & 0\\
 & \texttt{score\_granularity} & 60.7 & 16.7 & 91.7 & 0\\
 & \texttt{extreme\_level} & 47.2 & 0.0 & 100.0 & 2\\
 & \texttt{option\_closure} & 46.0 & 0.0 & 83.3 & 0\\
MEA.marginalisation & \texttt{score\_cumulative} & 1.4 & 0.0 & 4.2 & 0\\
 & \texttt{choice\_indicator} & 7.9 & 0.0 & 29.2 & 0\\
 & \texttt{pair\_choice\_marginals} & 35.5 & 0.0 & 75.0 & 0\\
 & \texttt{inclusion\_exclusion} & 46.0 & 0.0 & 79.2 & 0\\
LOG.bounds & \texttt{frechet\_and\_upper} & 54.8 & 25.0 & 87.5 & 0\\
 & \texttt{nary\_and\_upper} & 35.5 & 4.2 & 66.7 & 0\\
 & \texttt{nary\_and\_lower} & 98.6 & 91.7 & 100.0 & 15\\
 & \texttt{nary\_or\_upper} & 98.0 & 91.7 & 100.0 & 14\\
 & \texttt{nary\_or\_lower} & 51.6 & 4.2 & 95.8 & 0\\
 & \texttt{count\_upper} & 78.0 & 45.8 & 100.0 & 3\\
 & \texttt{count\_lower} & 97.2 & 70.8 & 100.0 & 15\\
LOG.monotonicity & \texttt{threshold\_sweep} & 61.5 & 0.0 & 95.8 & 0\\
 & \texttt{entailment\_strength} & 83.5 & 16.7 & 100.0 & 4\\
 & \texttt{quantifier\_monotonicity} & 92.1 & 33.3 & 100.0 & 9\\
CHO.conditioning & \texttt{irrelevant\_option} & 86.9 & 25.0 & 100.0 & 5\\
 & \texttt{remove\_option\_iia} & 72.4 & 45.8 & 100.0 & 4\\
 & \texttt{pairwise\_iia} & 67.3 & 4.2 & 100.0 & 4\\
\end{longtable}

}

\section{Menus}
\label{app:menu}

A Choice's output is a distribution given that the answer is one of its
options (\S\ref{sec:setting}). Three relations compare it with answers
that may not be conditioned on the menu: \texttt{choice\_indicator} and
\texttt{choice\_complement} ask a Noul about one option without listing
the others, and \texttt{option\_closure} replaces an option by ``none of
the other options''. The laws assume that the answer lies in the menu,
as the domain rules make it whenever the text determines the answer.
Under the alternative interpretation, the model uses one probability
measure $\mu$. A Choice assigns option $o$ the conditional probability
$\mu(o)/\mu(U)$, where $U$ is the union of the options. A Noul about
$o$ receives the unconditional probability $\mu(o)$. Then the Nouls of a menu are proportional to
the Choice and sum to $\mu(U)\le1$; ``not $o$'' gets
$1-\mu(o)\ge1-P_q(o)$; and a catch-all, which also covers the answers
outside the menu, gets at least the option it replaces.
Table~\ref{tab:menu} counts failures across all models that this
interpretation cannot explain at the same tolerance. These include
Noul probabilities whose sum exceeds $1+\varepsilon$, or whose
renormalised distribution differs from the Choice by more than
$\varepsilon$. They also include complements short of one by more
than $\varepsilon$ and catch-all probabilities more than
$\varepsilon$ below those of the replaced options. The reading explains few failures of the first two
relations; \texttt{option\_closure} is the one whose score it could
change.

\begin{table}[h]
\centering
\caption{Failures on JevBench-mini, over all models, that a menu not
containing the answer cannot explain.}
\label{tab:menu}
\small
\begin{tabular}{@{}lrrr@{}}
\toprule
Relation & Tests & Failed & Not explained\\
\midrule
\texttt{choice\_indicator} & \menuIndTests & \menuIndFailed & \menuIndUnexplained{} (\menuIndUnexplainedPct\%)\\
\texttt{choice\_complement} & \menuCompTests & \menuCompFailed & \menuCompUnexplained{} (\menuCompUnexplainedPct\%)\\
\texttt{option\_closure} & \menuCloTests & \menuCloFailed & \menuCloUnexplained{} (\menuCloUnexplainedPct\%)\\
\bottomrule
\end{tabular}
\end{table}

\section{Scores at other tolerances}
\label{app:tolerance}

\begin{table}[H]
\centering
\caption{Sensitivity of JevBench-mini scores to the tolerance. For each
tolerance, the table reports rank correlations with the ranking at
$\varepsilon=0.05$, the largest rank change, and the scores averaged
across the \nModels{} models and for the uniform reference and the
top-ranked model.}
\label{tab:tolerance}
\small
\begin{tabular}{@{}rcccccc@{}}
\toprule
$\varepsilon$ & Spearman & Kendall & max rank shift & mean score & uniform & first\\
\midrule
0.001 & 0.738 & 0.600 & 14 & 41.4 & 89.4 & Open-Jev-9B\\
0.01 & 0.858 & 0.705 & 9 & 55.7 & 89.4 & Open-Jev-9B\\
0.02 & 0.926 & 0.810 & 6 & 62.0 & 89.4 & Open-Jev-9B\\
0.03 & 0.978 & 0.905 & 3 & 66.0 & 89.4 & Open-Jev-9B\\
0.05 & 1.000 & 1.000 & 0 & 71.7 & 89.4 & Open-Jev-9B\\
0.075 & 0.973 & 0.886 & 3 & 76.2 & 89.4 & Open-Jev-9B\\
0.1 & 0.953 & 0.838 & 4 & 79.6 & 89.4 & Open-Jev-9B\\
0.15 & 0.778 & 0.608 & 12 & 84.2 & 94.1 & JevAny-Gemma-4B (pointer)\\
0.2 & 0.599 & 0.448 & 15 & 87.2 & 95.3 & Jebadiah 9B v2\\
\bottomrule
\end{tabular}

\end{table}

\section{Runtime}
\label{app:runtime}

\begin{table}[H]
\centering
\caption{Runtime per request, measured on each model's first run, and
resulting total time on one replica for JevBench-mini
(\nMiniRequests{} requests) and JevBench-240 (\nPlanRequests{}
requests). Each server used one A100 40GB GPU unless marked
$^\dagger$, which indicates CPU (ONNX Runtime, 12 threads).
jevhome-ettin-1b ran JevBench-240 on a slower CPU node.}
\label{tab:runtime}
\small
\begin{tabular}{@{}lrrrr@{}}
\toprule
Model & Params & s / request & mini, min & 240, min\\
\midrule
Verdict 1.4 & 151M & 0.037 & 0.8 & 6\\
Laya (english) & 421M & 0.055 & 1.1 & 9\\
Laya (typed-decisions) & 421M & 0.056 & 1.2 & 9\\
CLM v0.1 & 8.2B + 20M & 0.066 & 1.4 & 8\\
Kev-4B & 4.7B & 0.083 & 1.7 & 10\\
OpenThai-SystemOne & 0.75B & 0.134 & 2.8 & 22\\
RSI-Jev v3.0 2B & 2.3B & 0.155 & 3.2 & 26\\
TinyJev-4B & 4.0B & 0.175 & 3.6 & 29\\
decider-2b v10 & 1.9B & 0.194 & 4.0 & 33\\
jevhome-E$^\dagger$ & 48M & 0.232 & 4.8 & 40\\
JevK5 v0.3 & 4.2B & 0.351 & 7.3 & 61\\
JevAny-Gemma-4B (pointer) & E4B & 0.393 & 8.2 & 68\\
decider-4b v2.1 & 4.2B & 0.416 & 8.7 & 71\\
Jebadiah 9B v2 & 9.65B & 0.429 & 8.9 & 74\\
OneJev-4B & 5.2B & 0.463 & 9.6 & 66\\
JevAny-Qwen-4B (pointer) & 4B & 0.511 & 10.6 & 89\\
JevAny-Qwen-4B (direct token) & 4B & 0.530 & 11.0 & 91\\
Von 1.3 & 395M & 1.188 & 24.7 & 195\\
Bespoke-Nimble-9B & 9.7B & 2.620 & 54.5 & 400\\
Open-Jev-9B & 9.7B & 4.147 & 86.2 & 773\\
jevhome-ettin-1b$^\dagger$ & 1.0B & 6.109 & 127.1 & 1518\\
\bottomrule
\end{tabular}

\end{table}

\end{document}